\documentclass[twoside,11pt]{article}
\usepackage{jair, theapa, rawfonts}

\usepackage{graphicx}%
\usepackage{multirow}%
\usepackage{amsmath,amssymb,amsfonts}%
\usepackage{amsthm}%
\usepackage{mathrsfs}%
\usepackage[title]{appendix}%
\usepackage{xcolor}%
\usepackage{textcomp}%
\usepackage{manyfoot}%
\usepackage{booktabs}%
\usepackage{algorithm}%
\usepackage{algorithmicx}%
\usepackage{algpseudocode}%
\usepackage{listings}%
\usepackage{booktabs}

\usepackage{comment}
 \usepackage{array}
\usepackage{url}
\usepackage{threeparttable}
\usepackage{footnote}
\usepackage{tablefootnote}
\makesavenoteenv{table*}
\makesavenoteenv{table}
\usepackage[utf8]{inputenc}
\usepackage[T1]{fontenc}
\makesavenoteenv{tabular}

\jairheading{1}{1993}{1-15}{6/91}{9/91}
\ShortHeadings{From Principles to Practice: A Deep Dive into AI Ethics and Regulations}
{Sun, Miao, Jiang, Ding, \& Zhang}
\firstpageno{25}

\begin{document}

\title{From Principles to Practice: A Deep Dive into AI Ethics and Regulations}

\author{\name Nan Sun \email nan.sun@unsw.edu.au \\
       \addr University of New South Wales, 37 Constitution Ave, Canberra, ACT 2612 Australia
       \AND
       \name Yuantian Miao (Corresponding Author) \email sky.miao@newcastle.edu.au \\
       \addr University of Newcastle, University Dr, Newcastle, NSW 2308 Australia
       \AND
       \name Hao Jiang \email jh072535@foxmail.com \\
       \addr Swinburne University of Technology, John St, Melbourne, VIC 3122 Australia
       \AND
       \name Ming Ding \email ming.ding@data61.csiro.au \\
       \addr Commonwealth Scientific and Industrial Research Organisation (CSIRO), Level 5/13 Garden St, Sydney, NSW, 2015, Australia
       \AND
       \name Jun Zhang \email junzhang@swin.edu.au \\
       \addr Swinburne University of Technology, John St, Melbourne, VIC 3122 Australia}


\maketitle

\begin{abstract}
In the rapidly evolving domain of Artificial Intelligence (AI), the complex interaction between innovation and regulation has become an emerging focus of our society. Despite tremendous advancements in AI's capabilities to excel in specific tasks and contribute to diverse sectors, establishing a high degree of trust in AI-generated outputs and decisions necessitates meticulous caution and continuous oversight. A broad spectrum of stakeholders, including governmental bodies, private sector corporations, academic institutions, and individuals, have launched significant initiatives. These efforts include developing ethical guidelines for AI and engaging in vibrant discussions on AI ethics, both among AI practitioners and within the broader society. This article thoroughly analyzes the ground-breaking AI regulatory framework proposed by the European Union. It delves into the fundamental ethical principles of safety, transparency, non-discrimination, traceability, and environmental sustainability for AI developments and deployments. Considering the technical efforts and strategies undertaken by academics and industry to uphold these principles, we explore the synergies and conflicts among the five ethical principles. Through this lens, work presents a forward-looking perspective on the future of AI regulations, advocating for a harmonized approach that safeguards societal values while encouraging technological advancement.
\end{abstract}

\section{Introduction}\label{introduction}
In today's world, 
where Artificial Intelligence (AI) is swiftly reshaping numerous facets of our daily life, 
the demand for robust and efficient AI regulations has become more pronounced.
Acknowledging this need, 
on October 30, 2023, 
U.S. President Joe Biden enacted an Executive Order, 
signifying a historical move towards tackling the complex issues raised by AI technologies \cite{Biden}. 
This action highlights an increasing recognition of the profound influence of AI on aspects such as safety and security, privacy, fairness and civil rights, 
as well as innovation and competition, 
calling for a forward-thinking and preemptive strategy in regulatory measures regarding AI advancements. 

\begin{figure*}[h]
\centering
  \includegraphics[width=1\linewidth]{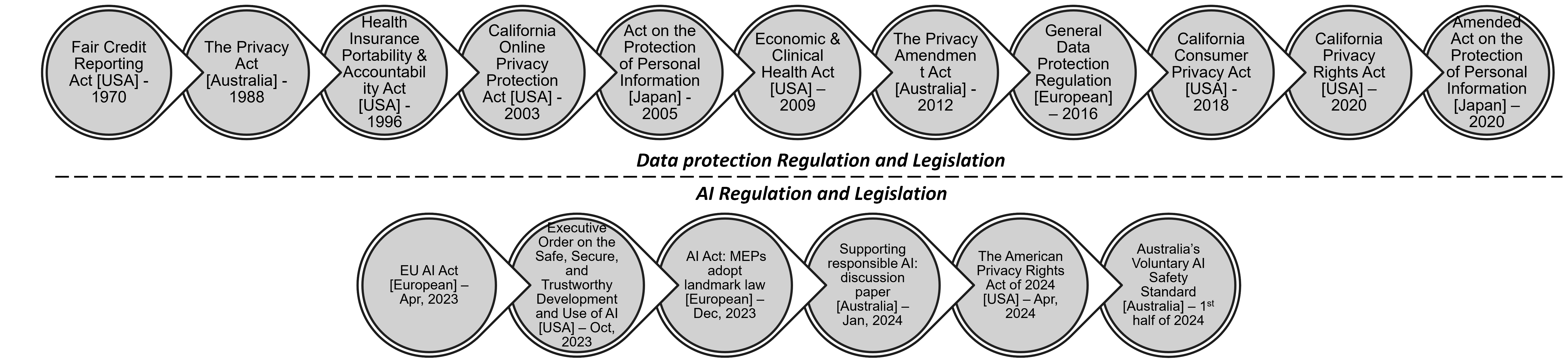}
  \caption{Timeline of data protection and AI regulations and legislation}
  \label{timeline}
\end{figure*}

The evolution of AI regulations represents a contemporary and dynamic narrative that has emerged significantly over recent decades. 
Initial apprehensions about ethics and data privacy have evolved into more defined guidelines and potential legislative frameworks, 
though the journey is far from complete. 
The chronicle of AI regulation is a modern and swiftly developing area, 
mirroring the rapid advancements in AI technology. 
The genesis of this regulatory trajectory can be traced back to the mid-20th century, 
marked notably by the introduction of the Fair Credit Reporting Act in 1970 in the United States \cite{fairCredit}. 
The legislation safeguards data gathered by consumer reporting agencies, 
including credit bureaus, medical data firms, and tenant screening services. 
The Act prohibits sharing information in a consumer report with individuals or entities that do not have a designated purpose as outlined in the Act. 
This act represented an early significant step in the formalization of data protection. 
Another significant achievement is the General Data Protection Regulation (GDPR) \cite{GDPR}, 
which is a thorough data privacy and security legislation enacted by the European Union (EU) in 2016. 
It stands as the most stringent data privacy and security law globally. 
While it was formulated and ratified by the EU, 
it places responsibilities on organizations worldwide 
as long as they interact with or gather data from individuals in the EU. 
The GDPR introduces a set of novel data privacy rights to grant individuals greater authority over the information they provide to organizations. 
As a visual representation, 
Figure \ref{timeline}, 
encapsulates the critical milestones in the progression of data protection regulations.


AI regulation goes beyond the data component to encompass AI systems' design, development, deployment, and use.
It is concerned not only with the data these systems process but also their decisions, the biases they may perpetuate, and the ethical implications of their interaction with humans and other systems.
AI regulation aims to address the societal and ethical challenges posed by increasingly autonomous technologies.
Regulation of AI varies from country to country, reflecting different approaches, priorities, and concerns. 
For example, in August 2023, the Australian Human Rights Commission (i.e., Commission) submitted a discussion document titled   ``Promoting Responsible AI'' to the Department of Industry, Science, and Resources in Australia \cite{AustraliaAIAct}. 
The Commission expresses particular concern about emerging issues, including privacy breaches, algorithmic discrimination, biases in automation, and the spread of misinformation and disinformation. 
Furthermore, the intricate interaction between AI, neurotechnology, the metaverse, and extended reality technologies adds complexity to this rapidly evolving field.
The Commission has advised the government to initiate an assessment of regulatory deficiencies to ascertain the relevance of existing legislation to AI.
In cases where gaps are identified, the associated legislation should be examined and updated to effectively address AI-related challenges.

Figure \ref{timeline} outlines the key developments in AI regulation and legislation across various regions, highlighting significant milestones from April 2021 to the first half of 2024.
In Europe, the EU AI Act, the first comprehensive regulation on AI, adopts a risk-based approach emphasizing safety, transparency, and environmental sustainability. It is expected to be fully implemented by 2026.
In October 2023, the White House issued an Executive Order to ensure safe, secure, and trustworthy AI development in the United States, prioritizing standards for AI safety and civil rights.
By April 2024, the U.S. introduced the American Privacy Rights Act, focusing on comprehensive data privacy.
Concurrently, Australia responded to AI safety discussions by proposing a voluntary AI safety standard emphasising a risk-based framework for AI deployment.
These initiatives reflect a growing international commitment to responsibly managing AI's societal impacts.


In the initial stages of research, 
Scherer \cite{scherer2015regulating} initiated a discussion on the feasibility and challenges of government AI regulation, suggesting paths for effective regulation.
Concurrently, other researchers delved into AI's regulatory and ethical aspects in specific domains. 
For example, Pesapane et al. \cite{pesapane2018artificial} explored these issues in medical services, while Reddy et al. \cite{reddy2020governance}  proposed a governance model for ethical and regulatory concerns in AI healthcare applications.
Focusing on particular aspects of AI regulation, W{\"a}schle \cite{waschle2022review} conducted a systematic literature review on highly automated driving, emphasizing safety assessment methods for AI systems. 
Other reviews have concentrated on different aspects: reviews \cite{larsson2020transparency,walmsley2021artificial} on AI transparency; works \cite{ferrer2021bias,caton2020fairness,pessach2022review}  on bias and discrimination in AI; and reviews \cite{van2021sustainable,nishant2020artificial,wu2022sustainable,dhar2020carbon} on AI's sustainability.

AI and algorithm-driven decision-making are increasingly impacting our daily lives, being extensively used in critical sectors, 
such as healthcare, business, government, education, and justice.
This shift towards an algorithmic society comes with many benefits and risks, as these systems can sometimes cause harm to users and society. Ensuring the safety, reliability, and trustworthiness of these systems is crucial. Trustworthy AI systems are those that are reliable, ethical, and transparent, enabling users to have confidence in the decisions made by these systems. Recent literature reviews on trustworthy AI \cite{li2023trustworthy,mora2021traceability,kaur2022trustworthy}, including work \cite{kaur2022trustworthy} focused on the elements of fairness, explainability, accountability, reliability, and acceptance to mitigate AI risks and enhance user and societal trust.
On the other hand, Li et al. \cite{li2023trustworthy} provided specific recommendations for AI practitioners, covering the entire AI system lifecycle, from data collection, model development, and system deployment to ongoing monitoring and governance.


In April 2021, 
the European Commission proposed the initial AI regulatory framework for the European Union \cite{AIAct}. 
This framework classifies AI systems into various risk levels, 
with corresponding levels of regulation. 
The primary objective of this Act is to guarantee the safety, transparency, traceability, non-discrimination, and environmental sustainability of AI systems used in the EU. 
The oversight of AI systems should rely on human decision-making (i.e., autonomy) rather than full automation to avoid adverse consequences. 
Members of the European Parliament endorsed the regulation, which was reached in negotiations with member states in December 2023, with 523 votes in favour, 46 against and 49 abstentions \cite{AIAct_currentstatus}.


To the best of our knowledge, 
there lacks a comprehensive survey that fully addresses the five fundamental principles laid out in the AI Act \cite{AIAct}, 
i.e., safety, transparency, traceability, non-discrimination, and environmental sustainability,
as they apply to AI models and systems. 
An analytical paper is needed to explore how regulations might balance the push for innovation with the need to mitigate risks such as unforeseen consequences or misuse of AI technology. 
In addition, 
a detailed review of existing AI regulations would enlighten policymakers, stakeholders, and the public and support more informed decisions in AI governance. 
Motivated by these considerations, 
we thoroughly examine AI regulation, 
exploring the interactions and tensions among its essential aspects. 
We also provide structured, actionable recommendations for AI professionals in academia and industry. The following is a summary of the main contributions of this paper.

\begin{itemize}
\item
We present and analyze the stipulations detailed in the AI Act, concentrating on the key areas covered by AI regulations, including safety, transparency, non-discrimination, traceability, and environmental sustainability within the realm of AI. These discussions are specifically rooted in the regulatory necessities and the foundational concepts introduced in the AI Act, providing a comprehensive overview of its scope and implications.
\item We discuss the synergies and conflicts among components of AI regulation through an analysis of current technical efforts focused on crafting AI systems that are safe, transparent, traceable, environmentally sustainable, and free from bias, respectively. This exploration aids in understanding how these elements interact and sometimes clash, guiding the development and deployment of AI that adheres to high ethical and regulatory standards.
\item By exploring the synergies and conflicts within these critical dimensions of AI regulation, we investigate how AI systems can be designed and developed to meet regulatory standards, highlighting the trade-offs involved. We also discuss future research pathways for scholars in AI and AI regulation fields and for industry practitioners who need to adhere to AI regulations once the AI Act is implemented.
\end{itemize}

\section{Preliminaries and definitions}
This section establishes the foundational concepts and definitions for this survey, beginning with a rationale for AI regulation to underscore its importance. We then introduce key concepts within the AI Act, emphasizing its focus on human-centric considerations.
\subsection{The need for AI regulation}
The journey of AI regulation is an evolving narrative, having gained significant momentum in recent years.
Initially, the focus was primarily on ethical considerations and data privacy concerns.
These foundational issues have gradually paved the way for more defined guidelines and, in some instances, the drafting of legislative proposals.
However, despite these strides, there is still a considerable amount of progress to be made in this area.
Before delving deeper into this topic, it's essential to outline the core reasons necessitating AI regulations, which range from safeguarding individual privacy and ensuring ethical AI use to mitigating potential societal risks and establishing accountability in AI development and application.

\subsubsection{Ethical and societal reasons}

The \textit{unbiasedness} of the AI system depends on its training data and design decisions. Without proper regulations, these systems risk perpetuating existing societal biases, leading to unfair or discriminatory outcomes. This is particularly concerning in areas such as employment, lending, and criminal justice, where biased AI could reinforce existing \textit{inequalities}. Regulations are necessary to ensure that AI systems are developed and deployed in a manner that is \textit{equitable} and \textit{fair} to all segments of society \cite{mehrabi2021survey}.

The increasing deployment of AI in critical sectors like healthcare, finance, and law enforcement raises important questions about \textit{accountability} \cite{bernard2023systematic}. When an AI system makes a decision that negatively impacts individuals or communities, it is essential to have clear guidelines determining who is \textit{responsible} – the developers, the users, or the AI itself. Regulations in this area would help clarify lines of responsibility and ensure that affected parties have recourse in the event of harm or injustice.

 AI's reliance on vast datasets, often comprising personal and sensitive information, heightens concerns about \textit{data privacy and security} \cite{chatterjee2023impact}. The potential for misuse or unauthorized access to this data poses significant risks to \textit{individuals' privacy rights}. Regulations are needed to establish strict guidelines on data collection, use, storage, and sharing. These include ensuring that individuals are informed and consented to how their data is used and implementing robust security measures to protect data from breaches and leaks.

Another reason is to mitigate the risk of \textit{social manipulation} \cite{sheikh2020understanding}. In particular, AI’s role in social media algorithms can influence public opinion and even manipulate elections, as evidenced by the Cambridge Analytica scandal, where personal data was used to target voters and potentially affect the political landscape. This has led to calls for tighter regulations on how AI can be used to manipulate information and target users on social media platforms through misinformation and disinformation.

\subsubsection{Technical reasons}

 In fields such as autonomous driving or medical diagnostics, where AI systems make decisions that can have life-or-death consequences, the importance of \textit{risk management} cannot be overstated \cite{ai2023artificial}. Regulations are critical in ensuring that these systems are designed, tested, and operated under stringent safety standards. These involve setting benchmarks for performance, establishing protocols for failure detection and response, and ensuring systems are resilient to various types of risks, including cyber threats, system malfunctions, and unexpected environmental conditions. Regulatory frameworks can guide the development and deployment of these systems to prevent accidents and mitigate potential harms.

From a technical standpoint, AI regulation is essential for promoting \textit{interoperability} and \textit{scalability}. Ensuring AI systems can effectively communicate and work together across different platforms and applications is crucial for maximizing their utility and efficiency. Regulations could establish standards that enhance this \textit{interoperability}. Additionally, \textit{scalability} is vital for AI systems to handle increasing loads smoothly. Regulations could help set guidelines that ensure AI systems are designed to scale efficiently without disproportionate increases in resource demands.

Furthermore, large AI models typically require substantial computational power, translating into significant energy consumption and consequent environmental impacts \cite{ahmad2021artificial}. Regulations could play a critical role in pushing for developing more \textit{energy-efficient AI technologies}, potentially reducing both the environmental footprint and the costs associated with high energy consumption.

\subsubsection{Global and National Security Reasons}


AI's capabilities in processing vast amounts of data make it a powerful tool for surveillance, espionage, and cyber warfare. This raises concerns about the protection of \textit{national interests and security} \cite{robinson2020trust}.
A regulatory framework is necessary to govern the use of AI in these domains, ensuring that it does not compromise national security or infringe on citizens' rights and freedoms.
Regulations can help define permissible uses of AI in surveillance and intelligence, establish safeguards against unauthorized data access, and protect against cyber threats that utilize AI technologies. This regulatory oversight is pivotal in maintaining the delicate balance between the use of AI for national security and the preservation of democratic values and individual rights.

The prospect of AI being used in \textit{autonomous weapons systems} is a growing concern \cite{akhtar2023regulating,de2020ethics}. These technologies can potentially revolutionize warfare, raising ethical and strategic questions about their deployment. Regulations are crucial in preventing the harmful use of AI in military applications, ensuring that such systems are developed and used in compliance with international humanitarian laws and ethical standards. By establishing clear boundaries and oversight mechanisms, regulations can prevent an arms race in lethal autonomous weapons and ensure that AI is used to enhance, rather than undermine global security and peace.

\subsubsection{Economic Reasons}

 AI technology has the potential to disrupt market dynamics significantly.
 Without proper regulatory oversight, there is a risk that large corporations with substantial resources could monopolize the AI sector.
 Such dominance could hinder innovation, as smaller companies and startups may find it challenging to compete on an unequal playing field. Regulations can foster \textit{a competitive market} by ensuring fair access to AI technologies and preventing monopolistic practices. This encourages a diverse ecosystem of AI developers and service providers, fostering innovation and ensuring that the benefits of AI are not confined to a few dominant players.

As AI becomes more integrated into consumer products and services, it is vital to \textit{protect consumers} from faulty or misleading AI applications.
Regulations are crucial in setting standards for AI quality, reliability, and truthfulness in advertising \cite{stojanovic2020can}. This includes ensuring that AI-driven products meet certain performance benchmarks and that claims made about AI capabilities are accurate and not misleading.
By doing so, regulations safeguard consumers from subpar or potentially harmful AI-driven products and services, fostering trust and confidence in the market.

\subsection{Key concepts in the AI Act}
\begin{figure*}[h]
  \centering
  \includegraphics[width=\linewidth]{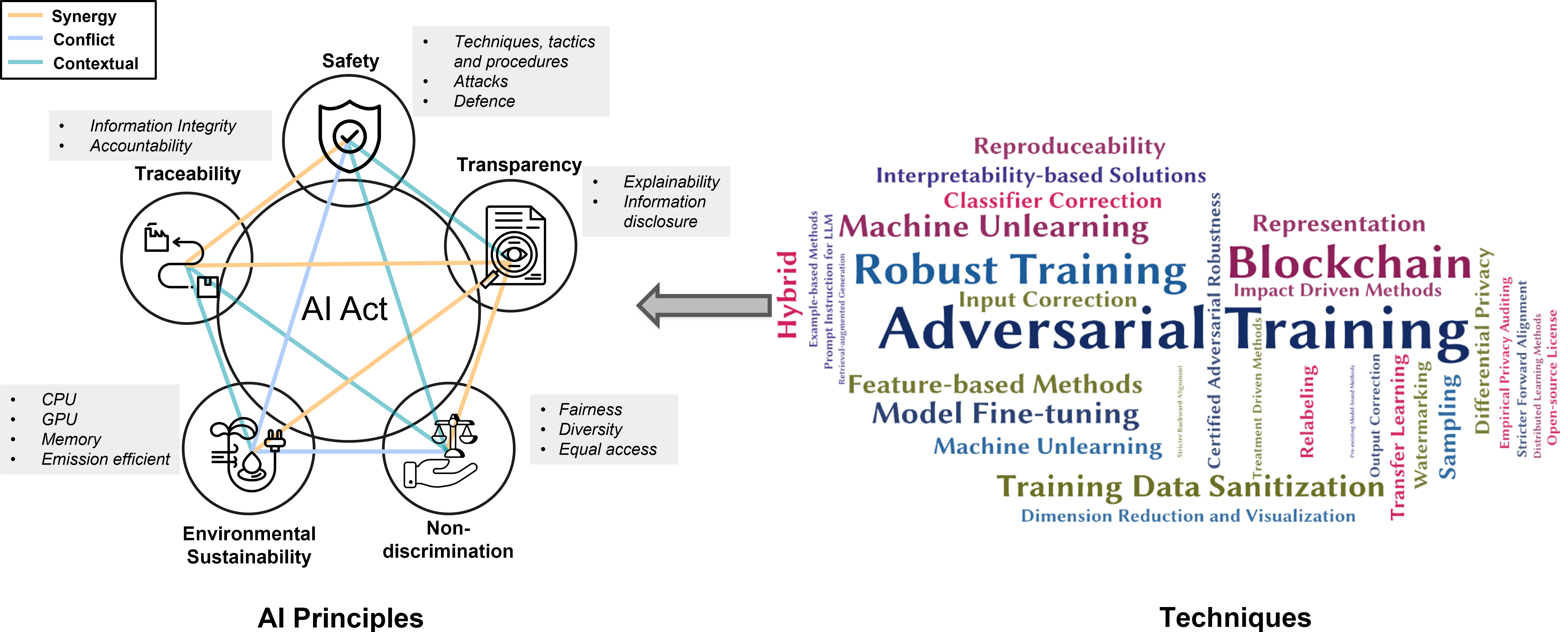}
  \caption{Human-centered principles of the AI Act and techniques supporting their implementation.}
  \label{definitions}
\end{figure*}
The AI Act aims to guarantee that AI systems utilized in the EU uphold safety, transparency, traceability, non-discrimination, and environmental sustainability, as illustrated in Figure \ref{definitions}. It places a strong focus on human oversight to avert detrimental consequences. In this context, we explain these five crucial elements as outlined in the AI Act.

\subsubsection{Safety}
AI safety, a multifaceted and interdisciplinary domain, focuses on preventing adverse consequences that might arise from AI systems \cite{leslie2019understanding}. 
This field tackles not only ethical considerations, ensuring AI systems align with moral values and contribute positively, but also addresses technical challenges \cite{du2021paradoxes}. 
These challenges include monitoring and maintaining the reliability of AI systems to avert risks.

In the EU AI Act, safety refers primarily to the regulation and mitigation of risks associated with using AI systems, ensuring that these systems do not pose unacceptable risks to health, safety, and fundamental rights \cite{Act_new}. Consider the machinery industry as an example.
High-risk AI systems here might involve AI-driven robots in manufacturing plants, such as an autonomous robotic arm assembling heavy machinery parts or an AI system managing critical manufacturing processes.
The malfunction of these systems can cause severe production setbacks, expensive damages, or even endanger human workers \cite{huang2021survey}. To mitigate these risks, it's crucial to equip AI systems with robust safety features like fail-safes or emergency stops. They must undergo comprehensive testing and validation across different operational scenarios.
Moreover, resilience against cyber threats or manipulation is vital to prevent potentially disastrous outcomes. 

\subsubsection{Transparency}

In the context of the EU AI Act, transparency refers primarily to obligations imposed on AI system providers to ensure that the operation and output of AI systems are clear and understandable by those who deploy or interact with them.
Transparency measures are designed to ensure that AI systems are used in a way that is understandable and respects the rights and freedoms of all individuals involved. This supports the overall goal of human-centric AI development and deployment within the EU.

The concept of transparency in AI regulation encompasses the level of openness and comprehensibility in how AI systems are developed, implemented, and managed \cite{robinson2020trust}. For instance, AI systems identified as high-risk must accompany these systems with detailed documentation and instructions that disclose the system's capabilities, characteristics, and limitations, allowing users to understand and correctly apply the outputs generated by these systems \cite{AIAct}. 

Beyond these operational aspects, transparency in AI also significantly intersects with the concepts of explainability\cite{hamon2020robustness}. This facet of transparency is about demystifying the internal workings of AI systems, providing clarity on how specific decisions are reached. It's a move towards making AI systems less of a ``black box'' and more of an open book, where users and stakeholders can understand and rationalize the logic behind AI-driven decisions. This level of transparency is not just a regulatory requirement but a foundation for building trust between users and AI technologies, ensuring that users can interact with these systems confidently and effectively, knowing the rationale behind their outputs and actions \cite{shneiderman2020bridging}.

\subsubsection{Non-discrimination}
The principle of non-discrimination in the realm of AI is centered on reducing the likelihood of biased decision-making by algorithms \cite{wachter2020bias}. This involves careful attention to the design and the integrity of data sets used in creating AI systems, combined with commitments to rigorous testing, effective risk management, thorough documentation, and consistent human oversight throughout the entire lifecycle of these AI systems.

In the context of AI regulations, non-discrimination is a guiding principle that ensures AI technologies are developed and utilized to prevent biased and unfair outcomes or decisions, particularly those that might unjustly target or disadvantage different groups of people\cite{AIAct}. This aspect of AI regulation is crucial given the risk that AI systems might unintentionally sustain, magnify, or even introduce new forms of biases or discriminatory practices \cite{lloyd2018bias}. Such regulations aim to encourage the creation and application of AI technologies in a manner that upholds fairness and ethical standards, thus preventing the reinforcement of existing social disparities. For instance, as outlined in OpenAI's GPT-4 technical report, content considered harmful encompasses elements like ``hate speech, discriminatory language, incitements to violence, or content that is then used to either spread false narratives or to exploit an individual'' \cite{OpenAIReport}. In response to this, GPT-4 was developed using a technique known as model-refusal. This approach, grounded in reinforcement learning, rewards the model during its training phase for actively declining to produce such harmful content.

\subsubsection{Environmental sustainability}

In AI regulation, the concept of environmental sustainability emphasizes the creation, implementation, and utilization of AI systems in ways that mitigate adverse environmental effects. This approach recognizes the ecological impact of AI technology, focusing on issues such as energy use, efficient resource management, and promoting ecological sustainability \cite{nishant2020artificial}. This includes safeguarding the environment and enhancing its quality, which also pertains to human health and safety \cite{AIAct}.

Crucial aspects of this environmental focus in AI regulation include promoting energy efficiency, particularly in AI systems with high computational demands like large data centers, which are significant electricity consumers \cite{ahmad2021artificial}. Additionally, it encompasses the pursuit of sustainable development, efforts to decrease the carbon footprint of AI technologies, and the advancement of `Green AI' \cite{van2021sustainable}. This latter term refers to designing AI models that are not just high-performing but also optimized for minimal consumption of computational resources and energy. Moreover, it involves adopting eco-friendly practices, such as minimizing electronic waste and encouraging the recycling of AI hardware components. 

\subsubsection{Traceability}

The principle of traceability within the framework of AI regulation encompasses the systematic capability to monitor, record, and chronicle the decision-making processes employed by AI systems during their entire operational lifespan \cite{AIAct}.
This principle is crucial for ensuring that AI operations are transparent and accountable. For instance, an AI system should possess robust logging features that provide a traceability level suitable for its intended use, as per the guidelines outlined in the AI Act. Such traceability measures are vital for verifying the AI system's performance and compliance throughout its lifecycle \cite{kroll2021outlining}.

Additionally, traceability involves closely tracking the various iterations of AI models, encompassing all modifications and updates that occur over time. This tracking is not just a record-keeping exercise but a critical tool for understanding how an AI model evolves \cite{agre2014toward}. It enables stakeholders to comprehend the changes made to the model, assess their impacts, and, if necessary, revert to prior versions when newer updates do not perform as expected or introduce unforeseen issues.
\begin{figure*}[h]
  \centering  \includegraphics[width=\linewidth]{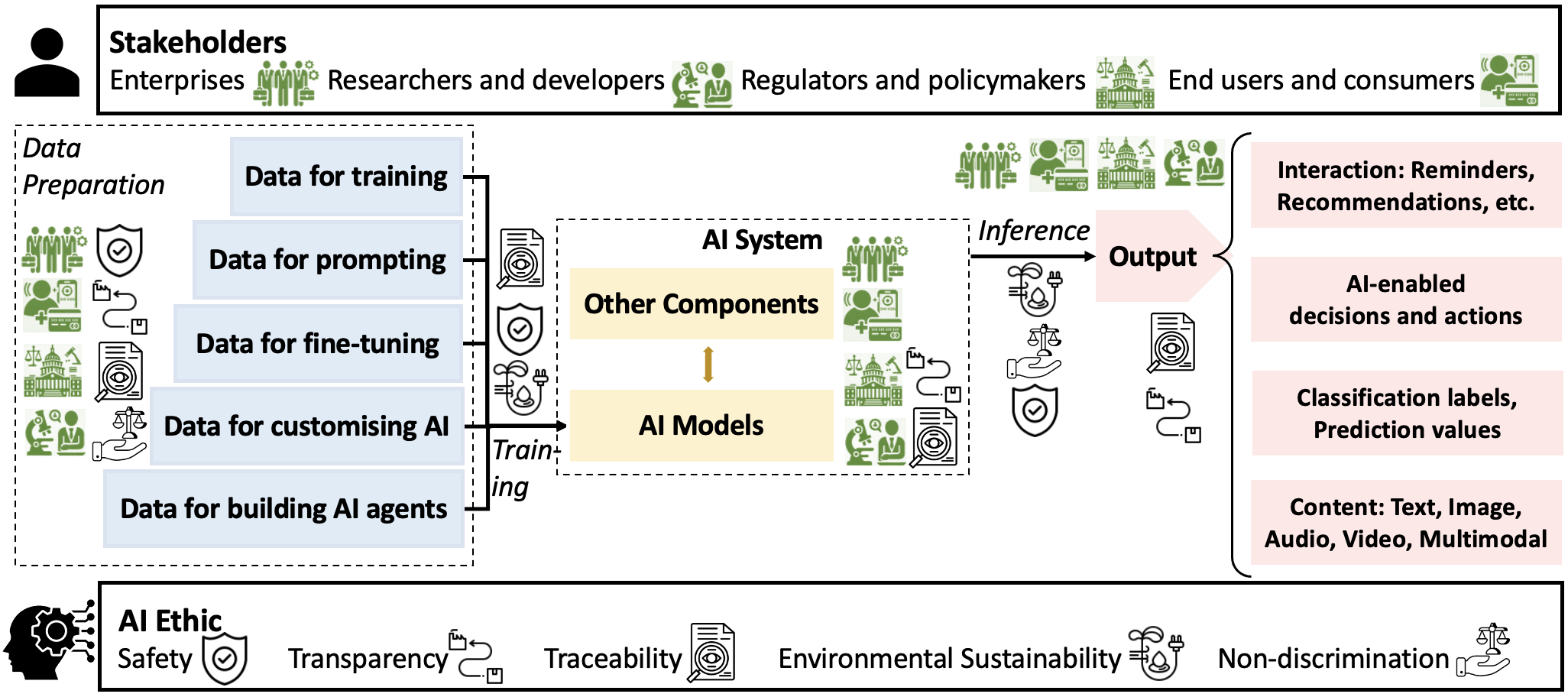}
  \caption{Human-centric principles reflected in the AI Act on the AI system components and the related stakeholders.}
  \label{fig:human_centric_ai}
\end{figure*}
\subsection{Human-centered AI Act}


The AI Act emphasizes putting people first. Hence, we discuss integrating human values into the design, development, and deployment of ethical AI systems in this subsection. As illustrated in Figure~\ref{fig:human_centric_ai}, we map various stakeholders' interests and concerns to the ethical considerations across each stage of the AI lifecycle. This comprehensive alignment ensures that AI systems are developed and deployed in a manner that addresses ethical aspects thoroughly, with the overarching goal of benefiting individuals and society at large.

We categorize stakeholders into four main groups: enterprises, researchers and developers, regulators and policymakers, and users and customers, as shown in Figure~\ref{fig:human_centric_ai}. According to recent research \cite{latonero2018governing,deshpande2022responsible,aguirre2020ai,stahl2022european}, these stakeholders have distinct interests and responsibilities regarding ethical AI development and deployment. 
Enterprises are responding to both internal and external pressures to ensure their AI technologies are ethically deployed and do not cause harm. 
For instance, Microsoft conducted a Human Rights Impact Assessment (HRIA) for AI, and Google has established AI principles that reference human rights \cite{latonero2018governing,deshpande2022responsible}. 
These efforts aim to mitigate adverse impacts, ensure corporate social responsibility, and maintain public trust.
Regulators and policymakers are responsible for aligning AI with human rights principles throughout each stage of the AI lifecycle.
Researchers and developers, who are deeply involved in each lifecycle stage, ensure technical performance and efficiency while promoting responsible AI behavior. 
They also contribute to the discourse on AI and human rights by examining the social impacts of AI, developing ethical frameworks, and advocating for policies that prioritize human dignity \cite{deshpande2022responsible}. 
Lastly, end-users and customers focus on the risks and harms associated with AI.
They advocate for AI systems that are free from biases, transparent in their operations and decision-making processes, particularly concerning conflicts of interest, and safe and traceable, including maintaining data integrity and security \cite{aguirre2020ai}.

The AI Act influences people by creating a comprehensive framework that addresses various aspects of AI development and deployment, all with the ultimate goal of protecting human safety, rights, and well-being while promoting beneficial AI technologies. It enhances safety by reducing risks of harm from AI applications, supports environmental sustainability to benefit both the environment and human well-being, ensures traceability for accountability, and increases transparency to foster trust, fairness, and informed decision-making. 

Ethical concerns in the data preparation stage include ensuring data safety, obtaining informed consent, and addressing biases in data collection \cite{huang2022overview}. It is crucial to collect representative data that accurately reflects the diversity of the population to avoid discrimination and ensure fairness. This process must also ensure the data owners' safety and make the data collection process transparent and traceable for data providers. During model training, it is essential to build a robust AI against various threats to ensure its safety, while the ethical use of computational resources should be considered to minimize environmental impacts. Transparency and traceability in the deployment process of AI models, including integrating other components within the AI system, are crucial to ensuring user safety. During the inference stage, it is important to evaluate the model's fairness, environmental sustainability, and safety in real-world scenarios \cite{huang2022overview}, and accordingly can opt to implement bias mitigation, cost-effective, and threats mitigation techniques. The output of the AI system is normally shown as various interactions with customers, AI-enabled decisions and actions, classification labels, prediction values, and/or other content like text, images, etc. All those output data must be transparent and traceable.




\section{How to design regulation-compliant systems: the synergies and conflicts}

This section investigates how the five key principles of AI regulation interact, exploring both their capacity to work together and the conflicts that exist between them. By analyzing these dynamics, this section aims to provide a thorough understanding of the techniques that can help AI systems adhere to these principles. This understanding will serve as a guide for developing AI systems that are both ethically sound and legally compliant.
\subsection{Safety}

\subsubsection{Overview}
\begin{figure*}[h]
  \centering
  \includegraphics[width=0.8\linewidth]{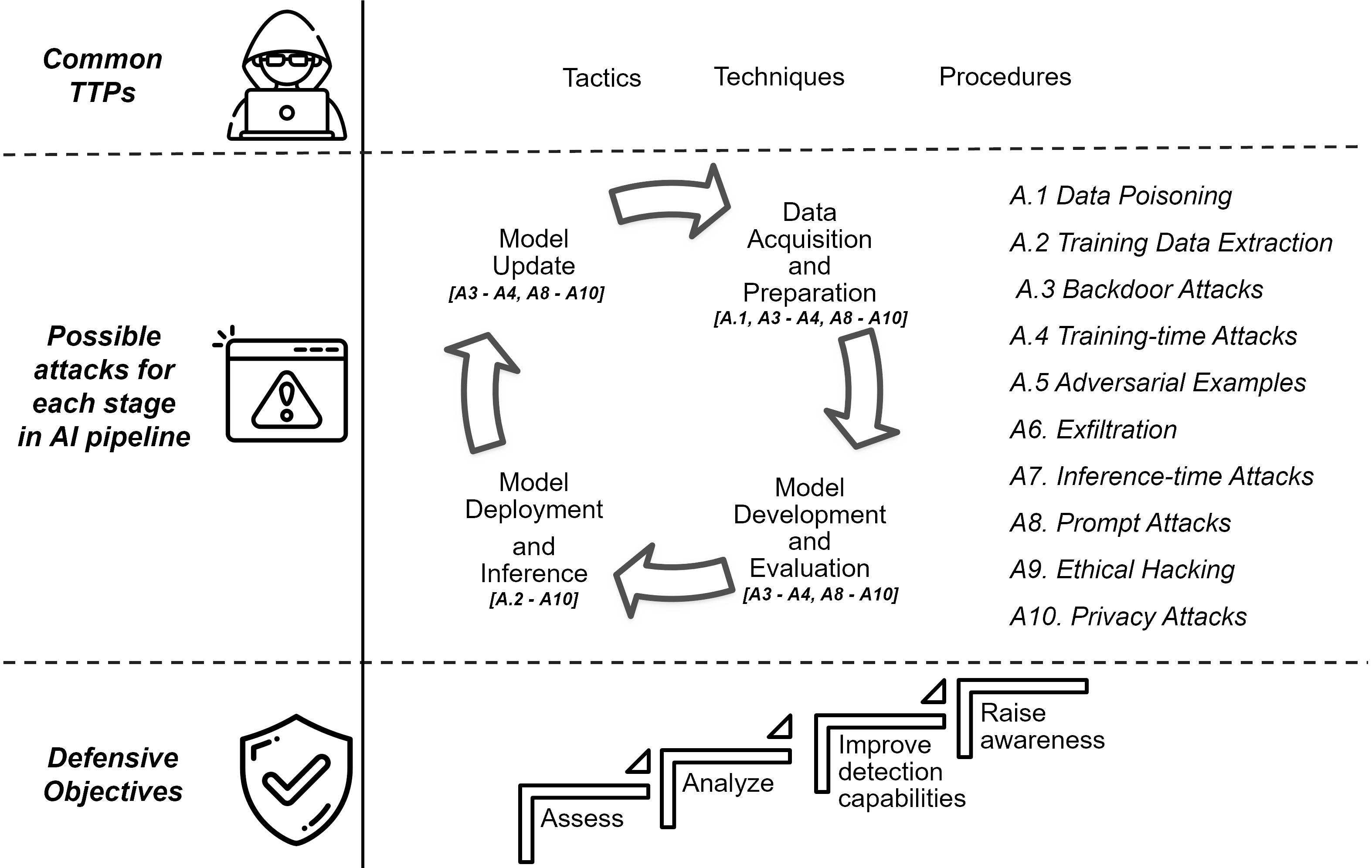}
  \caption{An overview of safe AI Systems.}
  \label{safety_overview}
\end{figure*}

Establishing well-defined safety protocols for the development and implementation of AI in an ethical manner is crucial.
Initially, we present a comprehensive overview of safety considerations in AI systems.
This forms the foundation for a more detailed exploration of how these safety considerations interact and potentially conflict with the other four key principles outlined in the AI Act: traceability, transparency, environmental sustainability, and non-discrimination.
As depicted in Figure \ref{safety_overview}, we identify potential attacks at each stage of the AI development pipeline. 
These attacks stem from the typical Tactics, Techniques, and Procedures (TTPs) employed by hackers.
Following this, we outline defensive objectives for the establishment of secure and reliable AI systems.

-\textit{Common TTPs:} Tactics, Techniques, and Procedures (TTPs) are integral concepts in cybersecurity, utilized to categorize and understand attacker behaviors. They serve as tools for testing and assessing the robustness of an organization's threat detection capabilities. Breaking down these concepts: (1) Tactics refer to the overarching goals or `why' behind an attack, explaining the motives for executing certain actions; (2) Techniques detail the `how' aspect, outlining the methods adversaries employ to achieve their tactical objectives; (3) Procedures are the specific, detailed methods used to execute these techniques.

There's a concerted effort within the AI security community to compile and comprehend the TTPs that adversaries might deploy against AI systems. A notable contribution in this domain comes from MITRE \cite{MITRE}, renowned for their MITRE ATT\&CK TTP framework. They have extended their expertise to the AI field by publishing a comprehensive set of TTPs tailored for AI systems \cite{MITREATLAS}. This work by MITRE is instrumental in aiding organizations to better understand and prepare for potential threats against their AI infrastructures, reflecting the growing importance of specialized TTPs in the evolving landscape of AI security.

-\textit{Possible attacks for each stage in AI pipeline:} Leveraging the knowledge in threat intelligence and AI system development, we have identified specific TTPs, leading to potential attacks that are most relevant and realistic for real-world adversaries. These potential attacks encompass a range of strategies, such as prompt attacks, extracting training data, backdoor attacks, adversarial attacks, data poisoning, exfiltration, and ethical hacking. The pipeline shown in Figure \ref{safety_overview} encompasses stages such as data collection and preparation, model creation and assessment, and the deployment and updating of models. This structured approach aids in comprehending specific vulnerabilities and developing appropriate countermeasures for each phase of AI system development, taking into account the safety considerations emphasized in the AI Act. We've organized possible attacks in a structured table as shown in Table \ref{Attacks_safety}, correlating each attack type with its respective phase in the AI development pipeline. In terms of their mechanisms and objectives, several of these attacks can overlap. For example, training-time attacks could encompass methods like data poisoning and backdoor attacks. Adversarial examples can be a form of both inference-time attacks and exfiltration if used to extract model behavior or data. Training data extraction and privacy attacks may involve unauthorized access to sensitive data. 

\begin{table*}[!ht]
\scriptsize
\caption{Potential Attacks Across Different Stages of the AI Pipeline 
Ensuring Safety}\label{Attacks_safety}
\begin{tabular}{p{1.5cm}p{6cm}p{1.5cm}p{5cm}}
\toprule
\textbf{Potential Attacks} & \textbf{Explanation} & \textbf{High risk stages} & \textbf{Related and latest surveys/ papers} \\
\midrule\midrule
Data Poisoning & Deliberate manipulation of training data to corrupt model learning, causing incorrect predictions. Common methods include injecting malicious data or altering existing data points. & Data Acquisition \& Preparation & \cite{cina2023wild}\cite{wang2022threats} \\
\midrule
Training Data Extraction & Unauthorized extraction or reconstruction of sensitive information from the training dataset used in model development. & Model Deployment \& Inference & \cite{carlini2021extracting} \\
\midrule
Training-time Attacks in GenAI & Attacks targeting pre-training and fine-tuning stages of GenAI models by manipulating portions of training data to cause specific model failures. & All stages & \cite{vassilev2024adversarial} \\
\midrule
Backdoor Attack & Insertion of hidden triggers into AI models that cause incorrect behavior only when specific inputs are present, while maintaining normal performance otherwise. & All stages & \cite{gao2020backdoor}\cite{li2023backdoor} \\
\midrule
Adversarial Examples Attack & Specially crafted inputs that cause AI models to produce incorrect outputs despite appearing normal to humans. Includes white-box, black-box, and transferability attacks. & Model Deployment \& Inference & \cite{chakraborty2021survey}\cite{akhtar2021advances}\cite{zhang2020adversarial} \\
\midrule
Exfiltration & Theft of sensitive data, proprietary algorithms, or intellectual property from AI models through unauthorized access or exploitation. & Model Deployment \& Inference & \cite{taori2023alpaca}\cite{chung2023implementing} \\
\midrule
Inference-time attacks in GenAI & Exploitation of vulnerabilities in deployed LLMs and RAG applications during the model's operational phase. & Model Deployment \& Inference & \cite{dong2024attacks} \\
\midrule
Prompt Attacks & Manipulation of LLM inputs through carefully crafted prompts to exploit model vulnerabilities and induce unintended behaviors. & All stages & \cite{shi2022promptattack} \\
\midrule
Privacy Attacks & Unauthorized extraction of sensitive information through techniques like membership inference, model extraction, and property inference attacks. & All stages & \cite{rigaki2023survey} \\
\midrule
Ethical Hacking & Authorized security testing of AI systems to identify and address vulnerabilities before malicious exploitation. & All stages & \cite{yaacoub2021survey} \\
\bottomrule
\end{tabular}
\end{table*}
-\textit{Defensive objectives:}
To enhance the safety of AI systems, we summarize four strategic objectives to defend against threat actors and attacks, inspired by considering the TTPs used by adversaries targeting various stages of the AI pipeline. The first objective involves evaluating the impact of attacks on both users and products, and determining methods to bolster cyber resilience against such incursions. The second objective focuses on examining the AI system's robustness, particularly in terms of detecting and thwarting potential attacks, as well as understanding how these attacks might circumvent existing safeguards. The third objective is to enhance the system's ability to detect threats. Finally, the fourth objective is to cultivate awareness among stakeholders. This is aimed at assisting AI system developers in recognizing risks and potential attacks at each stage of the AI pipeline, and in advocating for a risk-driven, well-informed approach to investing in security measures for AI systems for organizations.

\subsubsection{Techniques related to safety and their relations to other aspects of the AI Act}\label{subsubsec:safety_impacts}

Below, we explore methods to improve the security of AI systems in response to the attacks outlined in Table \ref{Attacks_safety}. Additionally, we examine how each method impacts the four principles outlined in the AI Act: transparency, non-discrimination, environmental sustainability, and traceability. 

\textbf{Adversarial training:} Adversarial training, introduced by Goodfellow et al. \cite{hamon2020robustness} and further developed by Madry et al. \cite{madry2017towards}, enhances model safety by incorporating adversarial examples during the training process. This method improves resilience against attacks and increases the model's semantic content \cite{tsipras2018robustness}, thereby boosting interpretability and predictability. These features are crucial for safety, allowing better anticipation and mitigation of potential failures. However, adversarial training often comes with trade-offs, including reduced accuracy on clean data and increased computational costs.

Formal verification employs methods from formal methods, such as Satisfiability Modulo Theories (SMT) solvers and abstract interpretation, to certify the robustness of neural networks against adversarial inputs \cite{katz2017reluplex} \cite{gehr2018ai2}. This mathematically rigorous approach enhances transparency by providing a solid foundation for understanding and documenting model behavior under adversarial conditions. While formal verification currently faces limitations in scalability and computational efficiency, it offers significant potential for comprehensive robustness certification. This rigorous verification is essential for developing transparent AI systems, as it allows for a deeper understanding of how and why a model is robust against certain types of adversarial inputs.

\textbf{Certified adversarial robustness:}
Certified adversarial robustness is crucial for AI system safety and transparency, ensuring predictable and reliable behavior under adversarial inputs. Randomized smoothing enhances classifier robustness by integrating Gaussian noise into input data \cite{lecuyer2019certified}\cite{cohen2019certified}, generating predictions most likely correct under noise perturbations. This approach yields provably robust classifiers, particularly against $\ell_2$-norm attacks \cite{carlini2021extracting}, enhancing system predictability and transparency. The method offers certified robustness for a subset of testing samples, providing quantifiable metrics to track model performance under adversarial conditions. Recent advancements have extended this concept to more complex perturbations, integrating denoising diffusion probabilistic models with high-accuracy classifiers, broadening the scope of transparency across various adversarial scenarios \cite{carlini2021extracting}.

\textbf{Training data sanitization:}
Training data sanitization enhances AI system safety by mitigating poisoned samples' impact. These methods identify and remove malicious samples before training. Techniques like the Region of Non-Interest (RONI) method enhance model reliability by excluding poisoned samples \cite{nelson2008exploiting}. Label cleaning methods target label flipping attacks \cite{paudice2019label}, while outlier detection \cite{steinhardt2017certified} and clustering methods \cite{laishram2016curie}\cite{taheri2020defending} remove anomalies. Ensemble variance computation helps identify potential poisoning attempts \cite{venkatesan2021poisoning}. Cybersecurity mechanisms for provenance and integrity attestation add further protection. Overall, data sanitization improves AI transparency by preserving training data authenticity, enhancing system safety and decision-making reliability.
\textbf{Robust training:}
Robust training enhances the transparency and safety of AI systems by integrating techniques like ensemble model voting, robust optimization, and randomized smoothing \cite{levine2020deep}\cite{wang2022improved}. Ensemble models allow for comparative analysis of individual model decisions, improving insight into the decision-making process. Robust optimization, through methods like trimmed loss functions, minimizes the impact of extreme data points, making the model's reasoning more transparency and less prone to data poisoning \cite{jagielski2018manipulating}\cite{diakonikolas2019sever}. Randomized smoothing adds stability to model predictions against minor input perturbations, offering a clearer understanding of how inputs lead to outputs and safeguarding against subtle adversarial attacks \cite{rosenfeld2020certified}. Collectively, these methods make AI systems' decision-making processes more discernible and reliable, fortifying them against manipulative data inputs.

\begin{table*}[!htbp]
\scriptsize
\caption{Mechanism/approaches influencing AI safety and the discussion on safety versus other four aspects
(``-'': negative impact, ``+'': positive impact).}
\begin{tabular}{p{4.65cm}p{1.35cm}p{2cm}p{5cm}p{2cm}}
\toprule
\textbf{Papers} & \textbf{Approaches Influencing AI Safety} & \textbf{Mitigations towards Potential Attacks} & \textbf{Impact} & 
\textbf{Other Aspects' Impacts in AI Regulation}\\
\midrule\midrule
\cite{goodfellow2014explaining,madry2017towards,tsipras2018robustness} & Adversarial training & Adversarial examples & The outcomes of adversarial training yield models with greater semantic significance compared to conventional models. & Transparency (+) \\
\midrule
\cite{lecuyer2019certified,cohen2019certified,katz2017reluplex,gehr2018ai2} & Certified robustness & Adversarial examples & By providing a clear understanding of how the model processes data and makes decisions, thereby allowing for the creation of robustness guarantees and the identification of model vulnerabilities. & Transparency (+) \\
\midrule
\cite{paudice2019label,steinhardt2017certified,taheri2020defending,venkatesan2021poisoning} & Training data sanitization & Data poisoning, backdoor & Before initiating machine learning training, ensure the training set is purified by eliminating any tainted samples. & Transparency (+) \\
\midrule
\cite{wang2022improved,jagielski2018manipulating,diakonikolas2019sever,rosenfeld2020certified} & Robust training & Data poisoning, backdoor & Adjusting the machine learning training algorithm to conduct resilient training rather than the standard approach. & Transparency (+) \\
\midrule
\cite{kairouz2021practical,chaudhari2023snap,mahloujifar2022property} & Differential privacy & Privacy attacks, training data extraction & Enhancing privacy protections for sensitive data used in AI training, testing, and deployment processes. & Fairness (+), transparency (+/-), traceability (+) \\
\midrule
\cite{jagielski2020auditing,zanella2023bayesian,nasr2021adversary} & Empirical privacy auditing & Privacy attacks, training data extraction & Empirically assessing an algorithm's privacy guarantees and establishing lower bounds through privacy attacks. & Fairness (+), transparency (+/-), traceability (+) \\
\midrule
\cite{bourtoule2021machine,izzo2021approximate} & Machine unlearning & Privacy attacks, training data extraction & Enhancing user privacy by ensuring that all data that a user wishes to be forgotten from a model is effectively removed. & Environmental-sustainability (-), fairness (+) \\
\midrule
\cite{ji2023ai,greshake2023not} & Stricter forward alignment & Prompt attacks & Developing integrated mechanisms through training for enhanced forward alignment and continuously refining via reinforcement learning informed by human feedback. & Fairness (+/-), transparency (+) \\
\midrule
\cite{liu2023prompt} & Prompt instruction & Prompt attacks & Prompting the model to handle user input with caution & Fairness (+/-) \\
\midrule
\cite{ji2023ai,liu2023prompt} & Stricter backward alignment & Prompt attacks & Developing built-in safeguards by enhancing backward alignment through training, using tailored benchmark datasets or filters to oversee the input and output of a secure LLM. & Transparency (+) \\
\midrule
\cite{kaur2022trustworthy,ruhr2023intelligent,OpenAIReport} & Interpretability-based solution & Prompt attacks & Making the models' decision-making processes more comprehensible to humans & Transparency (+) \\
\bottomrule
\end{tabular}
\end{table*}

\textbf{Machine unlearning:}
Machine unlearning represents a novel and distinctive approach aimed at mitigating privacy concerns in machine learning by allowing users to request the removal of their data from trained models, thus potentially enhancing privacy, fairness, and trust in AI systems. This technique comes in two primary forms: exact unlearning, which involves retraining the model from scratch or from a certain checkpoint to ensure complete removal of the data's influence, and approximate unlearning, where the model's parameters are adjusted to diminish the influence of the data intended to be forgotten \cite{bourtoule2021machine}\cite{cao2015towards}. Although promising, the implementation of machine unlearning, whether exact or approximate, introduces complexities related to ensuring the integrity and performance of the model post-unlearning, alongside considerations of computational efficiency and the potential environmental impact of the retraining or updating process \cite{ginart2019making}, \cite{izzo2021approximate}, 
\cite{neel2021descent}. As such, while machine unlearning is a forward step towards user-centric, privacy-preserving AI, it necessitates meticulous design and management to balance its benefits against the operational and ethical challenges it poses.

\textbf{Differential privacy:}
Differential Privacy (DP) enhances AI system safety by protecting sensitive training data \cite{dwork2006differential}. It limits an attacker's ability to gain insights about individual data points from AI outputs. DP has evolved to include approximate DP and Rényi DP \cite{mironov2019r}. The primary algorithm for training machine learning models is DP-SGD \cite{abadi2016deep}, with recent enhancements like DP-FTRL \cite{kairouz2021practical} and DP matrix factorization \cite{demontis2019adversarial}. DP protects against data reconstruction and membership inference attacks, with tight bounds derived by Thudi et al. \cite{thudi2022bounding}. However, it doesn't fully protect against model extraction or property inference attacks \cite{chaudhari2023snap}\cite{mahloujifar2022property}.

DP contributes to fairness by ensuring equal privacy protection across demographic groups \cite{farrand2020neither}. It enhances transparency in data handling practices \cite{dwork2006calibrating,gong2022transparent} and improves traceability of data usage. However, trade-offs between privacy and utility must be carefully considered in specific applications.

\textbf{Empirical privacy auditing:}
Implementing Differential Privacy (DP) faces challenges in balancing privacy and utility \cite{ponomareva2023dp}. Worst-case analysis often leads to higher privacy parameters in practice, as seen in the 2020 U.S. Census \cite{vassilev2024adversarial}. Privacy auditing, initiated by Jagielski et al. \cite{jagielski2020auditing}, aims to assess actual privacy levels through threat simulations. While membership inference attacks \cite{jayaraman2019evaluating}\cite{zanella2023bayesian} are used, poisoning attacks are more effective at revealing privacy erosion \cite{jagielski2020auditing}\cite{nasr2021adversary}.

Recent advancements include improved precision for the Gaussian mechanism \cite{nasr2023tight} and techniques reducing large sample size requirements \cite{pillutla2023unleashing}. Efficient privacy auditing approaches have been introduced by Steinke et al. \cite{steinke2023privacy} using random data canaries, and Andrew et al. \cite{andrew2023one} employing random client canaries for federated learning privacy evaluation.

\textbf{Stricter forward alignment:}
Prompt injection happens when a user deliberately inputs text to modify the behavior of a Large Language Model (LLM), aiming to circumvent safeguards for producing misinformation, offensive content, or extracting private information \cite{vassilev2024adversarial}. To counteract these attacks, model developers are implementing stronger built-in protections by enhancing forward alignment through training on selected, pre-aligned datasets and further refining the models via Reinforcement Learning informed by Human Feedback (RLHF) \cite{greshake2023not}\cite{ji2023ai}. This method of training, which indirectly incorporates human judgment to fine-tune models, helps ensure that LLMs are more aligned with human values and less prone to undesirable outputs. OpenAI's GPT-4, for instance, underwent fine-tuning with RLHF, resulting in a reduced propensity for generating harmful or inaccurate content \cite{OpenAIReport}. 

\textbf{Prompt instruction for LLM:}
LLM instructions can guide the model in handling user inputs with caution. By adding detailed instructions to the input prompt, models can be prepped to recognize and respond appropriately to content that might lead to a jailbreak, where the model's behavior deviates from intended safeguards \cite{chang2024play}. Strategically placing user input before these instructions can leverage the model's tendency to prioritize recent instructions, enhancing adherence to safety protocols \cite{liu2023prompt}. Furthermore, surrounding the instructions with random characters or specific HTML tags can signal to the model the distinction between system directives and user-provided prompts. This method helps delineate the boundaries between what the model perceives as commands to follow versus input to analyze, thereby improving the model's ability to navigate and process inputs more securely and effectively. Well-designed prompt instructions that prioritize fairness, sensitivity to context, and inclusive language can contribute to increased fairness in LLMs. However, it's essential to carefully consider the design and implementation of prompt instructions to avoid unintended consequences that could decrease fairness.

\textbf{Stricter backward alignment:}
Model providers are increasingly developing and integrating robust mechanisms by focusing on enhanced backward alignment. This involves rigorous training and evaluation processes using benchmark datasets specifically designed for this purpose or through filters that carefully monitor both the inputs received and outputs generated by a secured LLM. A notable strategy involves assessing the responses of LLMs to various prompts, which can help in identifying prompts that are potentially adversarial in nature, thereby preventing misuse of the model \cite{ji2023ai}.

Moreover, the emergence of commercial solutions offering tools to detect and mitigate prompt injection attacks represents a significant advancement in safeguarding LLMs, such as Arthur \footnote{https://www.arthur.ai/product/shield}, Lakera \footnote{https://www.lakera.ai/} and Aporia \footnote{https://www.aporia.com/}. These tools are designed to spot potentially harmful user inputs and moderate outputs to prevent the model from engaging in jailbreak behavior, where it acts outside its intended operational boundaries. By implementing such measures, companies are embracing a defense-in-depth approach, layering multiple security mechanisms to provide comprehensive protection for LLMs. This multifaceted strategy not only enhances the immediate security of these models but also contributes to a broader assurance of safety and transparency in their deployment and use.

\textbf{Interpretability-based solutions:} Interpretability-based solutions targeting attacks on LLMs involve using outlier detection techniques to monitor the models' prediction trajectories. This approach is grounded in the observation that how a model processes and predicts outcomes based on inputs can reveal a lot about the nature of those inputs. Specifically, researchers have found that by closely examining the ``prediction trajectory'' - the series of predictions an LLM makes as it processes input-it's possible to identify inputs that are anomalous or potentially malicious \cite{greshake2023not}.

The concept of a ``tuned lens'' refers to a method or tool designed to scrutinize the model's prediction path. When this lens is applied to analyze how an LLM responds to various inputs, it can highlight when the model's predictions deviate significantly from what is expected under normal circumstances. Anomalous inputs, such as those designed to manipulate the model or trigger undesired behavior, often lead to unusual prediction trajectories. By detecting these outliers, interpretability-based solutions can effectively flag potentially harmful inputs before they impact the model's output, enhancing the security and reliability of LLMs in handling a wide range of data \cite{belrose2023eliciting}.

\begin{table*}[!ht]
\caption{Comparison of transparency regarding information disclosure in leading AI tools}
\centering
\scriptsize
\begin{threeparttable}
\begin{tabular}{p{1.5cm}p{2cm}p{1.5cm}p{1.5cm}p{6.5cm}}
\hline
\textbf{\begin{tabular}[c]{@{}p{2cm}@{}}Organi-\\zation\end{tabular}} & 
\textbf{\begin{tabular}[c]{@{}p{0.5cm}@{}}Model\end{tabular}} & 
\textbf{\begin{tabular}[c]{@{}p{2cm}@{}}Official\\ model \\card or\\ technical report\end{tabular}} & 
\textbf{\begin{tabular}[c]{@{}p{2cm}@{}}API/\\SDK \\provided\end{tabular}} & 
\textbf{\begin{tabular}[c]{@{}p{6.5cm}@{}}Other details\end{tabular}} \\ 
\hline \hline
Meta & 
LLaMA 3\footnote{https://llama.meta.com/} & 
Y & 
Y & 
\begin{tabular}[c]{@{}p{6.5cm}@{}}LaMA 3, available in 8B and 70B configurations, is trained on publicly available data. The official model card discloses some details about the model architecture and training.\end{tabular} \\ 
\hline
OpenAI & 
ChatGPT 4\footnote{https://openai.com/chatgpt/} & 
Y & 
Y & 
\begin{tabular}[c]{@{}p{6.5cm}@{}}ChatGPT 4 is now accessible with the option to download model weights and a technical report. Although the overall structure and instructions for use are provided, detailed information about the training process is not revealed.\end{tabular} \\ 
\hline
Anthropic & 
Claude 3\footnote{https://claude.ai/} & 
Y & 
Y & 
\begin{tabular}[c]{@{}p{6.5cm}@{}}Claude 3 has unveiled its model weights and a technical report. The official documentation outlines details about the model training, yet specifics regarding the data sources, training process, and architecture are somewhat less detailed.\end{tabular} \\ 
\hline
Microsoft & 
Copilot\footnote{https://copilot.microsoft.com/} & 
Y & 
Y & 
\begin{tabular}[c]{@{}p{6.5cm}@{}}Copilot comes with official documentation and an open API. However, despite these resources, detailed information about the model's architecture and specific training details are kept under wraps.\end{tabular} \\ 
\hline
xAI & 
Grok 1.5\footnote{https://x.ai/} & 
N & 
N & 
\begin{tabular}[c]{@{}p{6.5cm}@{}}Due to the recent release of Grok1.5, detailed information about the model is currently scarce, permitting only preliminary usage via web interfaces.\end{tabular} \\ 
\hline
Google & 
Gemini 1.5 Pro\footnote{https://deepmind.google/technologies/gemini/pro/} & 
Y & 
Y & 
\begin{tabular}[c]{@{}p{6.5cm}@{}}Gemini 1.5 Pro has unveiled the fundamental information along with comparative test outcomes of the model. However, the details provided are still lacking for a comprehensive grasp of the architecture and training methods.\end{tabular} \\ 
\hline
\end{tabular}
\end{threeparttable}
\label{LLM_informationdisclosureD}
\end{table*}
\subsection{Transparency}
\subsubsection{Overview} 
AI transparency is crucial for upholding the integrity and comprehensibility of AI systems \cite{balasubramaniam2023transparency,cao2023extrapolation}. It involves straightforward communication about the functions, decision-making processes, and constraints of these systems. This level of openness enables users to understand how decisions are made and help build trust. Furthermore, AI transparency includes explainability, which requires providing clear explanations of how AI processes inputs to produce outputs. This aspect is essential for enhancing user confidence and ensuring that AI operations comply with regulations designed to protect rights and prevent biases. In this subsection, we explore transparency in AI by discussing the roles of explainability and information disclosure.

-\textit{Towards transparency through explainability:} AI explainability is crucial for understanding how AI models make decisions, ensuring transparency, accountability, and trust in AI systems. Several techniques can help achieve AI transparency by providing clear, understandable explanations for AI model behavior. Regarding \textit{model interpretability methods}, a model that is interpretable, such as a decision tree, a linear regression, or a rule-based system, is transparent, and the model itself is intrinsically understandable \cite{carvalho2019machine}. Techniques such as the SHapley Additive exPlanations (SHAP) and Local Interpretable Model-agnostic Explanations (LIME) offer insights into which features are most important for the model's predictions \cite{meza2023does}. These \textit{feature importance techniques} can help trace back the decision-making process by highlighting the contribution of each input feature to the final decision. Some works utilized methods, such as counterfactual explanations \cite{cheng2020dece} and prototypes \cite{van2021interpretable}, providing insights by presenting specific examples the model uses for making decisions. Specifically, counterfactual explanations show how slight changes to the input can lead to different predictions, helping trace the decision boundaries of the model. Prototypes, on the other hand, identify representative examples from the dataset that are influential in the model’s learning. In addition, \textit{visualizations} can offer intuitive insights into complex models, especially deep learning models \cite{samek2017explainable}. Techniques like saliency maps, which highlight the parts of the input (e.g., pixels in an image) that are most influential for a prediction, can help trace the model's focus and reasoning process \cite{ghariba2019visual}. Techniques that \textit{decompose model predictions} into understandable components. For instance, Layer-wise Relevance Propagation (LRP) decomposes the output decision back to the input features to identify the relevance of each feature in the decision-making process \cite{lapuschkin2019opening}. Lastly, tools and platforms allow users to interact with AI models, query them with different inputs and observe the outputs \cite{sun2022near}. This hands-on approach can help users understand the model's behavior in different scenarios and trace its decision-making process more practically and intuitively.

-\textit{Explainability for Large Language Models:} The techniques for enhancing Large Language Models (LLMs) explainability are categorized into two main paradigms: traditional fine-tuning methods and prompting-based methods \cite{zhao2023explainability}. Traditional fine-tuning methods encompass various strategies to enhance the interpretability and understanding of LLMs. \textit{Feature attribution methods} form one facet of this approach, discerning the significance of different parts of input data, such as words or tokens within a sentence, for the model's predictions. Techniques like Integrated Gradients \cite{enguehard2023sequential} utilize gradients to gauge the importance of each input feature by tracing their impact on the model's output. \textit{Surrogate models} offer another avenue, employing simpler and more interpretable models to approximate the intricate behavior of complex LLMs \cite{chen2022adversarial}. While they provide insights into decision-making processes, they may not capture the full complexity of the original model. \textit{Representation analysis} delves into understanding the learned representations of the model, often through visualization or clustering of embeddings to identify patterns or similarities captured by the model.

Prompting-based methods leverage the generative capabilities of LLMs for explanation generation \cite{zhao2023explainability}. By prompting the model to articulate reasons for its decisions, this method sheds light on its \textit{underlying reasoning}. \textit{Counterfactual explanations} involve altering inputs slightly to observe changes in the model's predictions, elucidating crucial aspects of the decision-making process \cite{treviso2023crest}. \textit{Prompt engineering} involves crafting prompts to elicit more interpretable or explainable responses from the model. These prompts may encourage the model to articulate its reasoning process or highlight key factors influencing its decisions, contributing to a deeper understanding of the model's behavior \cite{zou2023representation}.

-\textit{Information disclosure:}
Advances in generative AI have led to AI-generated media that is nearly indistinguishable from human-created content. This has underscored the need for standardized accountability reports as a valuable measure to improve the transparency of AI systems across the industry \cite{ali2024transparent}. According to the GDPR \cite{GDPR}, providers of AI systems interacting with humans must ensure that people are informed of their interaction with AI unless it's obvious from the context. Moreover, AI systems generating or manipulating content resembling real people, objects, or events (i.e., deepfakes) should explicitly disclose that the content is artificially created or manipulated.


A summary of detailed information concerning information disclosure in prominent AI systems is presented in Table \ref{LLM_informationdisclosureD}. The extent of disclosed information regarding the training dataset, model architecture, test outcomes, and human interaction varies at the present stage. Although advanced AI tools like GPT-4 have published detailed technical reports, OpenAI has withheld specific information on training and datasets, citing the need to protect product security and maintain competitiveness. In contrast, Meta's LLaMA 3 is considered relatively transparent, providing details on architecture, parameter count, and some training procedure information in its Model Card. Despite this, current leading AI tools remain insufficiently transparent regarding automated data processing and automated decision-making. Moreover, while most state-of-the-art AI tools offer corresponding APIs or SDKs to integrate them into local programming environments, understanding the intricacies of these models is challenging for researchers and users due to their complexity and encapsulation. As a result, achieving transparency in AI systems remains an elusive goal. 


\subsubsection{Transparency versus safety, Environmental sustainability, non-discrimination, and traceability}\label{subsubsec:trans_impacts}
While transparency strengthens traceability and supports non-discrimination by uncovering biases, it can conflict with safety and privacy concerns. Additionally, it promotes environmental sustainability but may reveal operational details. Thus, the ensuing discussion will elucidate the delicate balance required between transparency and the other principles to navigate the trade-offs and harmonize ethical AI deployment. As summarized in Table \ref{tab:Trans_impact}, the methods to augment transparency are categorized \cite{dwivedi2023explainable} into five categories: Dimension Reduction and Visualization, Feature-based Techniques, Example-based Techniques, Distributed Learning, and Pre-existing Model-based Methods. This subsection explores and discusses in more detail the approaches that can influence AI transparency under the six categories shown in Table \ref{tab:Trans_impact}.


\begin{table*}[!ht]
\caption{Mechanism/approaches influencing AI transparency and the discussion on transparency versus safety, non-discrimination, environmental sustainability, and traceability (``-'': negative impact, ``+'': positive impact).}
\label{tab:Trans_impact}
\centering
\scriptsize
\begin{tabular}{p{5cm}p{1.7cm}p{1.5cm}p{3.8cm}p{2.5cm}} 
\toprule
\textbf{Papers} & \textbf{Category} & \textbf{Methods} & \textbf{Impact} & \textbf{\begin{tabular}[c]{@{}c@{}}Discussion with\\ the Rest Aspects\\ in AI Regulation\end{tabular}} \\ 
\midrule \midrule
\begin{tabular}[c]{@{}p{5cm}@{}}\cite{nakanishi2024pcaldi,tharwat2017linear}\end{tabular} & 
\multirow{2}{*}{\begin{tabular}[c]{@{}p{1.7cm}@{}}Dimension reduction \& visualization\end{tabular}} & \begin{tabular}[c]{@{}p{1.5cm}@{}}PCA, LDA \end{tabular} & 
\multirow{2}{*}{\begin{tabular}[c]{@{}p{3.8cm}@{}}Visualizes feature distributions and their influence on model decisions, helping to understand model behavior and detect potential biases.\end{tabular}} & 
\multirow{2}{*}{\begin{tabular}[c]{@{}p{2.5cm}@{}}Traceability (+), non-discrimination (+), environmental sustainability (+).\end{tabular}} \\[5pt]
\cmidrule(r){1-1} \cmidrule(l){3-3}
\begin{tabular}[c]{@{}p{5cm}@{}}\cite{van2008visualizing,mcinnes2018umap,wexler2019if}\end{tabular} & 
& 
\begin{tabular}[c]{@{}p{1.5cm}@{}}t-SNE, UMAP What-If\end{tabular} & 
& 
\\[8pt] \midrule
\begin{tabular}[c]{@{}p{5cm}@{}}\cite{goldstein2015peeking,ribeiro2018anchors,lundberg2017unified,nori2019interpretml}\end{tabular} & 
\begin{tabular}[c]{@{}p{1.7cm}@{}}Feature based methods\end{tabular} & 
\begin{tabular}[c]{@{}p{1.5cm}@{}}ICE, LIME, SHAP, InterpretML\end{tabular} & 
{\begin{tabular}[c]{@{}p{3.8cm}@{}}Explains prediction outcomes by analyzing feature importance and their individual contributions to model decisions.\end{tabular}} & 
\begin{tabular}[c]{@{}p{2.5cm}@{}}Traceability (+), non-discrimination (+).\end{tabular} \\[5pt] \midrule
\begin{tabular}[c]{@{}p{5cm}@{}}\cite{dwivedi2023explainable,dhurandhar2018explanations,garcia2020shapley,mothilal2020explaining,fisher2019all,wexler2019if}\end{tabular} & 
\begin{tabular}[c]{@{}p{1.7cm}@{}}Example based methods\end{tabular} & 
\begin{tabular}[c]{@{}p{1.5cm}@{}}Anchors, CEM, KT SHAP, Dice, Alibi, What-If\end{tabular} & 
{\begin{tabular}[c]{@{}p{3.8cm}@{}}Demonstrates how specific changes in input conditions affect model predictions, making decision processes more interpretable.\end{tabular}} & 
\begin{tabular}[c]{@{}p{2.5cm}@{}}Traceability (+), non-discrimination (+).\end{tabular} \\[8pt] \midrule
\begin{tabular}[c]{@{}p{5cm}@{}}\cite{warnat2021swarm,kairouz2021advances}\end{tabular} & 
\begin{tabular}[c]{@{}p{1.7cm}@{}}Distributed learning methods\end{tabular} & 
\begin{tabular}[c]{@{}p{1.5cm}@{}}Federated Learning, Swam Learning\end{tabular} & 
{\begin{tabular}[c]{@{}p{3.8cm}@{}}Enables transparent model behavior adjustment at each network node while maintaining data privacy through limited sharing.\end{tabular}} & 
\begin{tabular}[c]{@{}p{2.5cm}@{}}Safety (-+), non-discrimination (+), environmental sustainability (-).\end{tabular} \\[5pt] \midrule
\begin{tabular}[c]{@{}p{5cm}@{}}\cite{bourtoule2021machine}\end{tabular} & 
\begin{tabular}[c]{@{}p{1.7cm}@{}}Pre-existing model based methods\end{tabular} & 
\begin{tabular}[c]{@{}p{1.5cm}@{}}Machine Unlearning\end{tabular} & 
{\begin{tabular}[c]{@{}p{3.8cm}@{}}Updates model behavior by modifying features or removing specific data influences.\end{tabular}} & 
\begin{tabular}[c]{@{}p{2.5cm}@{}}Non-discrimination (+), environmental sustainability (+).\end{tabular} \\[5pt]
\bottomrule
\end{tabular}
\end{table*}

\textbf{Dimension reduction and visualization:} These techniques are essential for achieving data understanding and interpretability, thereby promoting transparent AI models. They simplify the complexity of large datasets, providing clearer insights and more transparent data narratives that can be integrated into AI algorithms for decision-making. For instance, Principal Component Analysis (PCA) \cite{shlens2014tutorial} and Linear Discriminant Analysis (LDA) \cite{tharwat2017linear} can reduce high-dimensional features to low-dimensional spaces for feature space visualization, roughly explaining a model's decisions. Although these transformations may obscure the original features' meanings, the importance of features in model decisions can still be prioritized based on their similarity to PCA loadings \cite{shlens2014tutorial}. Visualization techniques like t-SNE and UMAP offer a visual understanding of the model's decision boundaries and the distribution of the training data \cite{van2008visualizing,mcinnes2018umap,wexler2019if}.

These techniques also positively impact the development of non-discriminative and environmentally sustainable AI with enhanced traceability. Specifically, PCA and LDA can unintentionally amplify biases if the principal components or discriminants used for dimensionality reduction capture biases inherent in the data \cite{shlens2014tutorial,tharwat2017linear}. In contrast, t-SNE \cite{van2008visualizing} and UMAP \cite{mcinnes2018umap} preserve local structures and neighborhood relationships, helping to reveal and mitigate biases to ensure equitable treatment of all individuals and groups. Regarding environmental sustainability, dimensionality reduction with PCA and LDA can make model training more computationally efficient, leading to lower energy consumption and a smaller carbon footprint. Visualization techniques like t-SNE and UMAP can also reduce computational costs by providing insights that lead to more efficient model training and fewer iterations. Additionally, t-SNE and UMAP maintain relationships between data points in a lower-dimensional space, facilitating the traceability of how groups of data points are related to each other and potential biases, thus enhancing the traceability of decisions within the AI system.

\textbf{Feature-based methods:} Deployed during the Deployment and Maintenance stage, these techniques elucidate the contribution of individual features to the outcomes of AI models. They allow for the dissection of model decisions, facilitating an understanding of how inputs are transformed into outputs. Feature-based methods such as ICE (i.e., Individual Conditional Expectation) \cite{goldstein2015peeking}, LIME (i.e., Local Interpretable Model-Agnostic Explanations) \cite{ribeiro2018anchors}, SHAP (i.e., Shapley Additive explanations) \cite{lundberg2017unified}, and Interpret-ML \cite{nori2019interpretml} significantly influence factors like non-discrimination and traceability in the realm of AI transparency.

These methods enhance non-discrimination by providing granular insights into model predictions. ICE plots offer a microscopic view of how changing one feature affects the prediction, which can reveal discriminatory patterns against certain groups if predictions change unfavorably for specific feature ranges. LIME focuses on local fidelity, giving interpretable explanations for individual predictions, which can be crucial in pinpointing and addressing instances where the model might be unfairly biased against certain samples. SHAP values quantify the contribution of each feature to every prediction, providing a detailed breakdown that can uncover feature-related biases in model behavior. Together, these methods enable the identification and correction of discriminatory aspects within models, leading to more equitable AI systems. Meanwhile, ICE, LIME, and SHAP are especially effective in improving traceability within AI systems. ICE plots enable a clear understanding of the model's behavior with respect to individual features, providing traceable insights. LIME facilitates traceability by offering interpretable and locally accurate explanations for individual predictions, allowing one to follow the logic behind the model’s output. SHAP extends this by assigning each feature an importance value for each prediction, creating a transparent and traceable link between input features and their impacts on the output. This ability to trace back the decision-making process to specific model inputs and their interactions significantly enhances the transparency of AI systems.

\textbf{Example-based methods:} Throughout the deployment and maintenance stage, these techniques focus on providing explanations for individual predictions made by AI systems. They support transparency by enabling stakeholders to query and receive understandable explanations for specific AI decisions. Example-based methods such as Anchors \cite{dwivedi2023explainable}, Contrastive Explanation Method (CEM) \cite{dhurandhar2018explanations}, Kernel Tree SHAP \cite{garcia2020shapley}, Diverse counterfactual explanations (Dice) \cite{mothilal2020explaining}, and Alibi \cite{fisher2019all} positively impact non-discrimination and traceability in the context of AI system transparency:

These methods provide detailed explanations based on instances, which can be pivotal in identifying and addressing discrimination within AI systems. For instance, Anchors offer explanations in the form of rules that are sufficient to guarantee the same prediction in similar cases, which can highlight discriminatory patterns if certain groups are consistently associated with negative outcomes. Similarly, CEM can indicate if a model's predictions are adversely affected by certain demographics, and Dice provides counterfactual explanations that can demonstrate if and how outcomes could be equitable under different circumstances. By illustrating the specific conditions that lead to particular decisions, example-based methods enable stakeholders to detect and rectify biases, contributing to the non-discrimination of AI systems. Thus, traceability is significantly enhanced by example-based methods. They provide concrete instances that make the decision-making process of AI systems transparent and understandable. For example, Kernel and Tree SHAP calculate the contribution of each feature to a particular prediction, clarifying the decision-making process. Alibi offers diverse types of model explanations, including feature influence, which aids in tracing back the logic of the AI's conclusions. Dice creates counterfactual scenarios that can be traced to understand how different inputs would change the outcomes, thereby improving the system's transparency.

\textbf{Distributed learning methods:} 
Distributed learning methods offer decentralized AI solutions that align with inherently decentralized data structures and comply with data privacy and security regulations. These methods, like federated learning \cite{kairouz2021advances} and swarm learning \cite{warnat2021swarm}, ensure that raw data does not need to be exchanged, maintaining proper transparency for all participants. Specifically, swarm learning provides secure, transparent, and fair onboarding for decentralized network members without requiring a central custodian \cite{warnat2021swarm}.

While improving AI system transparency, distributed learning methods have mixed impacts: they positively affect non-discrimination, negatively affect environmental sustainability, and have complex effects on safety. Regarding non-discrimination, collaborative fairness can be achieved by adjusting the performance of models allocated to each participant based on their contributions \cite{lyu2020collaborative}. However, distributed learning methods are not environmentally friendly in AI development. Qiu et al. \cite{qiu2020can} found that training time for distributed learning is significantly longer than for centralized training due to the lower computational capabilities of participants' devices. The impact on safety is multifaceted. A survey highlights the positive impact on data privacy through distributed learning in AI development but also points out vulnerabilities to various attacks, such as poisoning, backdoor, Generative Adversarial Network (GAN)-based attacks, and inference-based attacks \cite{gosselin2022privacy}.

\textbf{Pre-existing model-based methods:} 
Unlike the aforementioned post-hoc explainable AI methods, this category encompasses approaches that ensure AI model transparency through modifications or adjustments to the model's capabilities. Specifically, machine unlearning \cite{bourtoule2021machine} ensures the user's ``right to be forgotten'' by enabling the model to erase the knowledge associated with a specific user. As discussed in section~\ref{subsubsec:safety_impacts}, this method also has a positive effect on non-discrimination but positively impacts the environmental cost of AI.

\subsection{Environmental Sustainability}
\subsubsection{Overview}
The concept of environmental sustainability in AI systems emphasizes a balance between the environment, society, and economy \cite{van2021sustainable}, ensuring that AI innovations contribute positively to each area without causing adverse effects on the environment. Researchers recognize the significant carbon footprint and energy demands of extensive IoT networks and cloud/edge communications, advocating for an architectural and deployment approach that minimizes environmental impact \cite{wu2022sustainable}. The environmental impact of training a single extensive machine learning model, like Meena, can be equated to driving an average passenger vehicle for 242,231 miles \cite{patterson2021carbon}. However, this represents merely one facet of the broader picture. The environmental cost not only exists in the AI model training process but also in the AI system life cycle. As depicted in Figure~\ref{safety_overview}, we explore potential environmental costs at each stage of the AI development pipeline. After that, we summarize the strategies applied to optimize the cost at each stage. Finally, we analyze the consequences of constructing an environmental sustainability AI system, focusing on its implications for safety, transparency, non-discrimination, and traceability.

The environmental costs mainly focus on power/energy consumption and carbon emissions caused by the usage of CPUs, GPUs, memory, and other system software and hardware. We specifically analyze those impact factors during data acquisition and preparation, model training, model inference, and AI system manufacturing and deployment. and summarize corresponding optimization strategies.

-\textit{Data Acquisition and Preparation:}
In this phase, energy consumption is predominantly driven by two processes: (1) the loading and validation split of large datasets, and (2) data preprocessing and feature extraction. The former necessitates downloading and storing training data on the model training server, a process that, although essential, is relatively less energy-intensive. Bouza et al. \cite{bouza2023estimate} observed that loading a 6 GB Imagenet validation split accounted for merely 0.5\% of the total energy consumption compared to model training. The latter process involves feature extraction and the weighting of individual features to assess the model's training efficacy, requiring significant computational resources. This stage is marked by exploring a wide array of machine-learning concepts on a large scale. Wu et al.~\cite{wu2022sustainable} reported that the energy expenditure for this process at Meta was half that of the model training phase.
\begin{table*}[!ht]
\caption{Environmental cost measurement in CPU, GPU, and memory consumptions and emission efficiency. ('UNK'=unknown, 'avg'=average)}
\label{tab:EF_tools}
\centering
\scriptsize
\renewcommand{\arraystretch}{1.2}
\begin{tabular}{>{\raggedright\arraybackslash}p{1cm}>{\raggedright\arraybackslash}p{3cm}>{\raggedright\arraybackslash}p{3cm}>{\raggedright\arraybackslash}p{3cm}>{\raggedright\arraybackslash}p{3.7cm}}
\hline
\textbf{Tools} & \textbf{CPU Consumption} & \textbf{GPU Consumption} & \textbf{Memory Consumption} & \textbf{Emission Efficient} \\ 
\hline \hline
Green-\begin{tabular}[p]{@{}p{1cm}@{}}Algori-\\thm\footnote{www.green-algorithms.org}\end{tabular} & 
Energy: avg 12W/core if UNK; TDP. Usage Factor: 100\% usage if UNK. & 
Energy: TDP; nvidia-smi; or avg 200W/GPU. Usage Factor: 100\% usage if UNK & 
Energy: 0.3725W/GB of memory available. Usage Factor: all available/requested memory. & 
Location: not restricted. Static data: Carbon Foot-print; Electricity Maps. Default: 475gCO2eq/kWh \\ 
\hline
Code-\begin{tabular}[c]{@{}c@{}}carbon\footnote{https://github.com/responsibleproblemsolving/energy-usage}\end{tabular} & 
Energy: RAPL files or Power Gadget; TDP. Usage Factor: avg value 50\% & 
Energy: pynvml library (NVIDIA GPUs only). Usage Factor: whole machine consumption & 
Energy: 0.3725W/GB of memory available. Usage Factor: allocated memory by 'process' & 
US \& Canada: regional emissions per unit of power consumed; Other Countries: avg energy mix from Global Petrol Prices. Default: 475gCO2eq/kWh \\ 
\hline
Eco2AI\footnote{https://github.com/sb-ai-lab/Eco2AI} & 
Energy: avg 100W/core if UNK; TDP. Usage Factor: uses os \& psutil python modules & 
Energy: pynvml library (NVIDIA GPUs only). Usage Factor: whole machine consumption & 
Energy: 0.3725W/GB of memory available. Usage Factor: allocated memory by 'process' & 
Each country's energy mix; Default: 436.5gCO2eq/kWh \\ 
\hline
Carbon-\begin{tabular}[c]{@{}c@{}}Tracker\footnote{https://github.com/lfwa/carbontracker}\end{tabular} & 
Energy: RAPL files. Usage Factor: whole machine RAPL value & 
Energy: pynvml library (NVIDIA GPUs only). Usage Factor: whole machine consumption & 
Energy: use RAPL files. Usage Factor: used by 'process' \& others & 
Real-time data: Energi Data Service (Denmark); Carbon Intensity API (Great Britain). Static data: carbon-intensities.csv. Default: 475gCO2eq/kWh \\ 
\hline
EIT\footnote{https://dl.acm.org/doi/abs/10.5555/3455716.3455964} & 
Energy: avg 100W/core if UNK; TDP. Usage Factor: uses os \& psutil python modules & 
Energy: nvidia-smi (NVIDIA GPUs only). Usage Factor: 'nvidia-smi -q -x' & 
Energy: use RAPL files. Usage Factor: used by 'process' \& others & 
Real-time data: California ISO. Static data: co2eq\_parameters.json Default: 301gCO2eq/kWh \\ 
\hline
MLCO2\footnote{https://mlco2.github.io/impact/} & 
None & 
Energy: TDP. Usage Factor: max load & 
None & 
Static data: the impact.csv file from cloud provider \\ 
\hline
Cumula-\begin{tabular}[c]{@{}c@{}}tor\footnote{https://github.com/EPFLiGHT/cumulator}\end{tabular} & 
Energy: avg 250W/core if UNK; TDP. Usage Factor: None & 
Energy: avg 250W/GPU if UNK; TDP. Usage Factor: max load & 
None & 
Static data: Electricity Maps. Default: 447gCO2eq/kWh \\ 
\hline
\end{tabular}
\end{table*}

-\textit{Model Training:}
Upon selecting an appropriate training solution, the AI model undergoes training using a more comprehensive set of production data, which is not only more current but also larger in volume and enriched with features. The carbon emissions associated with model training can be divided into two types: offline and online training emissions. Offline training leverages historical data for model training, whereas online training dynamically updates the model parameters using the latest data. Wu et al. noted that, at Meta, the carbon emissions from online training were significantly lower than those from offline training \cite{wu2022sustainable}.

Additionally, the selection of batch sizes and epoch configurations influences energy consumption. Selecting an optimal batch size represents a balance between energy consumption and runtime \cite{bouza2023estimate}, with larger batch sizes offering faster runtime and reduced energy usage. Nonetheless, excessively large batch sizes can lead to inefficient energy consumption due to low usage rates. Therefore, the ideal batch size is the maximum size that ensures full utilization of all GPUs and CPUs. Similarly, epoch configuration compromises the model's quality and runtime. The duration of epochs and their associated energy consumption (or carbon emissions) remain constant \cite{anthony2020carbontracker}. A larger number of epochs extends the runtime, while also enhancing the likelihood of achieving model convergence. The model structure configuration directly impacts the amount of the model's parameters. 

Beyond the previously mentioned factors, checkpointing exerts a negligible effect, and the impact of the number of parameters on energy use and carbon emissions is uncertain. Bouza et al.~\cite{bouza2023estimate} identified an inconsequential difference in energy consumption between experiments conducted with a single checkpoint versus ten checkpoints. Typically, a larger number of parameters suggests extensive computational efforts during training. Nevertheless, actual energy consumption varies based on the specific model setup and training infrastructure employed. For example, training a Switch Transformer model, which boasts 1.5 trillion parameters \cite{fedus2022switch}, results in considerably lower carbon emissions compared to the GPT-3 model, which contains 750 billion parameters \cite{mann2020language}.


-\textit{AI System Manufacturing and Deployment:}
The manufacturing and deployment of AI systems significantly affect carbon emissions and energy consumption. The impacts related to AI system manufacturing and deployment include the model's implementation program, the quantity and capabilities of processors running the program, the efficiency of data centers in power delivery and cooling, and the mix of energy sources (e.g., renewable, gas, coal) \cite{patterson2021carbon}. For instance, Meta's Transformer-based Universal Language Model for text translation at the inference stage served a lot of production traffic resulting in large energy consumption \cite{conneau2019unsupervised}. Bouza et al. \cite{bouza2023estimate} also highlight the impact of infrastructure size on power consumption through comparative experiments conducted on different platforms. For instance, training the Denoiser model on the larger Gemini-1 infrastructure (Grid5000 server) was completed in 2 hours, whereas it took 3 hours and 16 minutes on the smaller Rosenblatt infrastructure (MAP5 server). Furthermore, the CPU usage factor was significantly lower on Grid5000 (16\%) compared to MAP5 (39\%), and a similar trend was observed for GPU usage, with 14.3\% on Grid5000 and 54\% on MAP5, indicating higher efficiency in larger infrastructures. The energy-intensive nature of producing advanced computing components, coupled with the electricity demands of data centers hosting AI applications, underscores the environmental implications of AI's widespread adoption.

A denial-of-service (DoS) attack at the AI system deployment stage can generate sponge examples, significantly increasing energy consumption and the model's inference run time \cite{shumailov2021sponge}. This attack exploits the energy-latency gap vulnerability in AI systems, where different inputs of the same size can cause a DNN to consume varying amounts of time and energy. This disparity is due to optimization strategies in hardware and AI algorithms. For instance, custom and semi-custom hardware typically leverage data sparsity and low-precision computations for DNN inference, reducing both arithmetic complexity and DRAM traffic to achieve better power efficiency \cite{chen2019eyeriss,han2016eie,zhao2019automatic}. However, attackers can craft sponge examples by manipulating the token size of an NLP model's input and output sentences and the size of embedding spaces, leading to a substantial increase in algorithmic complexity and energy consumption. A proposed defense is to set a worst-case performance bound for some models and monitor their usage \cite{shumailov2021sponge}.


Table \ref{tab:EF_tools} summarized a list of tools \cite{bouza2023estimate} used to measure CPU, GPU, and memory consumption and the emission efficiency related to the AI system's manufacturing and deployment. Users can monitor their AI system's environmental cost in its development and deployment to determine which stage needs to be improved to achieve environmental sustainability.

-\textit{Environmental cost optimizations:}
In the realm of AI, environmental cost optimization encompasses several pivotal areas, including algorithmic enhancements, processor improvements, data center efficiency, and the strategic mix of energy sources. A notable advancement in algorithm development is illustrated by the Evolved Transformer (Medium) model, which through neural architecture search, operates with significantly lower computational demands-1.6 times fewer FLOPS and reduced processing time by 1.1 to 1.3 times compared to its predecessor, the Transformer (Big), while simultaneously achieving a marginal increase in accuracy \cite{patterson2021carbon}. This progress underscores the potential for algorithmic and program improvements to contribute directly to environmental sustainability by reducing the energy footprint of AI operations.

Further optimizations are evident in the development of specialized hardware for deep learning, aimed at enhancing the cost-performance ratio. The focus on Total Cost of Ownership (TCO), which includes operational costs like electricity consumption and the capital expenses associated with computing infrastructure, highlights the relationship between power usage and financial expenditure. Patterson et al.~\cite{patterson2021carbon} shows that enhancements in performance per TCO also yield benefits in performance per watt, leading to financial savings and a reduction in carbon emissions. Moreover, cloud data centers exhibit significantly higher energy efficiency compared to traditional enterprise data centers, attributed to factors such as improved server utilization. This efficiency has mitigated the anticipated increase in energy consumption attributed to data center operations. Additionally, the strategic deployment of cloud computing facilities in locations with cleaner energy grids or where clean energy can be directly procured (e.g., Finland and Iowa) leverages the efficiency of transmitting information through optical fibers over long distances, further diminishing the environmental impact of AI systems. This holistic approach to environmental cost optimization in AI system manufacturing and deployment highlights the synergy between technological advancements and strategic operational decisions in reducing the carbon footprint and energy consumption of AI technologies.

\subsubsection{Environment sustainability versus safety, transparency, non-discrimination, and traceability}\label{subsubsec:ef_impacts}
Apart from environmental cost optimization, some techniques are considered for building environmentally sustainable AI during the model development and deployment processes. We analyze whether those environmentally sustainable techniques could simultaneously satisfy the other four aspects of an ethical AI. Table~\ref{tab:EF_impacts} summarizes their impacts and corresponding discussion.

\begin{table*}[!htbp]
\caption{Mechanism/techniques influencing AI environmental sustainability and the discussion on environmental sustainability versus safety, transparency, non-discrimination and traceability (``-'': negative impact, ``+'': positive impact)}
\label{tab:EF_impacts}
\centering
\scriptsize
\begin{tabular}{p{3.5cm}p{1cm}p{4.5cm}p{3.2cm}p{2.5cm}}
\toprule
\textbf{Papers} & \textbf{Category} & \textbf{Techniques} & \textbf{Impact} & \textbf{Discussion with Other Aspects in AI Regulation} \\
\midrule \midrule

\parbox{4cm}{\cite{mazumder2023reg}} & 
\multirow{8}{*}{\parbox{1cm}{Model Fine-tuning}} & 
\parbox{4.5cm}{A regression-focused profiling: find the various metrics' trend with hardware deployment of NN} & 
\parbox{3.3cm}{Energy-efficient configuration for model development on the target device} & 
\multirow{6}{*}{\parbox{2.5cm}{Safety (+), Non-discrimination (+)}} \\[5pt]
\cmidrule(r){1-1} \cmidrule(l){3-4}

\parbox{3.5cm}{\cite{sha2022fine}} & & 
\parbox{4.5cm}{A super-fine-tuning apply a dynamic learning rate method} & 
\parbox{3.2cm}{Fast learning with regular change learning rate end in low energy cost} & \\[5pt]
\cmidrule(r){1-1} \cmidrule(l){3-4}

\begin{tabular}[c]{@{}p{3.5cm}@{}}\cite{labonte2024towards,kirichenko2022last}\end{tabular} & & 
Last-layer fine-tuning & & \\[5pt]
\midrule

\begin{tabular}[c]{@{}p{3.5cm}@{}}\cite{guo2019spottune,chen2021user,chen2019transfer}\end{tabular} & 
\multirow{14}{*}{\parbox{1cm}{Transfer Learning}} & 
\parbox{4.5cm}{Adaptive or layer-wise fine-tuning: select layer to be fine-tuned} & 
\multirow{7}{*}{\parbox{3.2cm}{Fewer layers fine-tuned reduce memory footprint \& computational costs}} & 
\multirow{10}{*}{\parbox{2.5cm}{Non-discrimination (+/-), transparency (+)}} \\[5pt]
\cmidrule(r){1-1} \cmidrule(l){3-3}

\begin{tabular}[c]{@{}p{3.5cm}@{}}\cite{zhao2011cross,nater2011transferring}\end{tabular} & & 
\parbox{4.5cm}{Parameter-based learning share parameters between models} & & \\[5pt]
\cmidrule(r){1-1} \cmidrule(l){3-3}

\begin{tabular}[c]{@{}p{3.5cm}@{}}\cite{mihalkova2008transfer,mihalkova2007mapping}\end{tabular} & & 
\parbox{4.5cm}{Relation-based learning establish effective model mapping} & & \\[5pt]
\cmidrule(r){1-1} \cmidrule(l){3-4}

\begin{tabular}[c]{@{}p{3.5cm}@{}}\cite{dai2007boosting,fang2019adapted}\end{tabular} & & 
\parbox{4.5cm}{Instance-based learning adapts data transferring knowledge} & 
\parbox{3.2cm}{Negative impact with training enlarged dataset} & \\[5pt]
\cmidrule(r){1-1} \cmidrule(l){3-4}

\parbox{3.5cm}{\cite{wu2022cost}} & & 
\parbox{4.5cm}{Phantom tree algorithm to measure the transfer cost} & 
\parbox{5cm}{Trade-off among memory, computational cost and model performance} & \\[5pt]
\midrule

\begin{tabular}[c]{@{}p{3.8cm}@{}}\cite{bourtoule2021machine,kurmanji2024towards}\end{tabular} & 
\multirow{4}{*}{\parbox{1cm}{Machine Unlearning}} & 
\parbox{4.5cm}{Exact unlearning with sub model/checkpoints retrained to entirely eliminate the data trace} & 
\multirow{4}{*}{\parbox{3.2cm}{Lower computational cost by avoiding the entire model retraining}} & 
\multirow{4}{*}{\parbox{2.5cm}{Non-discrimination (+), transparency (+)}} \\[5pt]
\cmidrule(r){1-1} \cmidrule(l){3-3}

\begin{tabular}[c]{@{}p{3.5cm}@{}}\cite{neel2021descent,izzo2021approximate}\end{tabular} & & 
\parbox{4.5cm}{Approximate unlearning with a gradient-based deletion approach to limit training data's impact} & & \\[5pt]
\midrule

\parbox{3.8cm}{\cite{lewis2020retrieval}} & 
\parbox{1cm}{RAG} & 
\parbox{4.5cm}{Pre-trained language models with information retrieval methods to elevate text creation} & 
\parbox{3.2cm}{Lower computational cost with the elimination of training expenses} & 
\parbox{2.5cm}{Non-discrimination (+)} \\[5pt]
\midrule

\begin{tabular}[c]{@{}p{4cm}@{}}\cite{guo2020multi,liu2021discrimination}\end{tabular} & 
\multirow{8}{*}{\parbox{1cm}{Model compression}} & 
\parbox{4.5cm}{Pruning: Reduce model parameters that contribute little in training} & 
\multirow{6}{*}{\parbox{3.2cm}{Memory- \& energy-efficiency in model deployment and/or development with reduced model size}} & 
\multirow{8}{*}{\parbox{2.5cm}{Non-discrimination (-), transparency (+/-), traceability (-)}} \\[5pt]
\cmidrule(r){1-1} \cmidrule(l){3-3}

\begin{tabular}[c]{@{}p{3.5cm}@{}}\cite{yamamoto2021learnable,gong2020vecq}\end{tabular} & & 
\parbox{4.5cm}{Quantization: Reduce the number of bits used to represent each weight} & & \\[5pt]
\cmidrule(r){1-1} \cmidrule(l){3-3}

\begin{tabular}[c]{@{}p{3.5cm}@{}}\cite{chen2021logarithmic,hsu2022language}\end{tabular} & & 
\parbox{4.5cm}{Low-rank factorization: Identify model's redundant parameters with matrix \& tensor decomposition} & & \\[5pt]
\cmidrule(r){1-1} \cmidrule(l){3-3}

\begin{tabular}[c]{@{}p{4cm}@{}}\cite{ji2021show,wang2021knowledge,huang2022knowledge}\end{tabular} & & 
\parbox{4.5cm}{Knowledge distillation: Transfer knowledge from a large to a smaller model \& ensure the validity} & & \\[5pt]
\bottomrule
\end{tabular}
\end{table*}

\textbf{Model fine-tuning:} Model fine-tuning stands as a cornerstone technique in machine learning, streamlining the enhancement of an existing model to better align with specific tasks or data sets. As a cornerstone of transfer learning, this method allows for the seamless transfer of knowledge from one domain to a related one, facilitating the swift customization of pre-trained models for new tasks with limited labeled data \cite{guo2019spottune}. This approach not only significantly boosts the accuracy of models for targeted applications but also underscores the efficiency of model fine-tuning in elevating overall model performance.

Model fine-tuning offers an eco-friendly alternative to the traditional, resource-intensive process of training a model from the ground up using extensive datasets \cite{chen2021user,labonte2024towards,kirichenko2022last}. By utilizing a smaller training dataset, fine-tuning significantly reduces computational and storage requirements during the Data Acquisition and Preparation phase. This reduction not only decreases the computational burden associated with loading, preprocessing, and extracting features from the data but also minimizes memory demands for data storage. In the Model Training phase, the smaller dataset and/or fewer layers to be trained facilitates quicker training times and manageable computational expenses. When applied within distributed or cloud-based systems, the minimized dataset size also leads to more manageable data communication costs, enhancing the overall efficiency and sustainability of the model development process. 

Additionally, this technique plays a pivotal role in creating environmentally sustainable AI by optimizing platform/application-level caching \cite{wu2022sustainable}. By leveraging pre-computed embeddings from existing models, fine-tuning improves power efficiency in developing new models for distinct tasks, thereby minimizing the environmental footprint of AI training processes. Mazumder et al.~\cite{mazumder2023reg} suggested a profiling approach to identify the most efficient fine-tuning configuration in terms of energy consumption or latency, targeting the required accuracy. This method analyzes the energy/latency contours across various tinyML device platforms.

Furthermore, model fine-tuning is instrumental in bolstering safety and fostering non-discrimination within AI development. Comprehensive fine-tuning can markedly diminish the threat of backdoor attacks, safeguarding the utility of AI systems while maintaining high safety standards 
. End-to-end fine-tuning effectively eradicates hidden vulnerabilities, showcasing its critical role in securing AI technologies. Moreover, adjusting the final layer of a model can enhance fairness, mitigating bias and promoting equality \cite{liang2023architectural,henzinger2023monitoring,mao2023last}. This harmonious integration of environmental sustainability, safety measures, and non-discriminatory practices through model fine-tuning exemplifies its comprehensive benefits in fostering the creation of ethical and responsible AI systems.

\textbf{Transfer learning:} Transfer learning is advocated for scenarios where training data is either scarce or costly to acquire, enabling the use of different domains, tasks, and distributions for training and testing purposes \cite{niu2020decade,pan2009survey}. This approach, akin to model fine-tuning, promotes the development of eco-friendly AI by reducing energy consumption during the data acquisition, preparation, and model training stages. Unlike model fine-tuning, which focuses on preserving the pre-learned features, transfer learning seeks to adapt the pre-trained model's weights, features, or architectural knowledge to a new task, requiring minimal further training \cite{wu2022cost}. In addition to benefits similar to fine-tuning, this discussion examines the environmental advantages of transfer learning through its methods of transferring knowledge.

To enhance model performance by facilitating the effective transfer of useful knowledge and preventing the transfer of detrimental knowledge during the Data Acquisition and Preparation phase. This effort is outlined in the work of Chen et al. \cite{chen2019transfer} and further detailed by Niu et al. \cite{fang2019adapted}, who categorize the methodologies into four distinct types: instance-based, feature-based, parameter-based, and relation-based learning. Notably, parameter-based and relation-based methods are highlighted for their potential to support the development of AI systems in an environmentally sustainable manner.

Instance-based and feature-based learning do not significantly contribute to the environmental sustainability of AI, while some specific methods may negatively impact it. Instance-based learning involves adapting data from the source domain for use in the target domain \cite{dai2007boosting,fang2019adapted}, thereby enlarging the training set, which typically leads to training the model from scratch without any environmental benefits. Feature-based learning, while seeking domain-invariant features, can lead to a negative environmental impact. This type of learning often requires feature mapping techniques that transfer large amounts of data from the source to the target domain and involve adversarial learning and deep learning to extract high-level features. The latter methods can significantly amplify the computational burden during data acquisition and preparation. Wu et al. \cite{wu2022cost} introduced an innovative and cost-effective transfer learning framework, OPERA (Online Transfer using Phantom Tree for Real-Time Adaptation), to optimize the trade-off between accuracy improvement and computational expense. This framework employs the phantom tree algorithm, which assesses the complexity involved in constructing tree-based models, as a method to evaluate the costs associated with transfer learning.

Conversely, learning about parameters and relations offers a more environmentally friendly approach. Parameter-based learning leverages shared parameters between models from the source and target domains, akin to model fine-tuning discussed above \cite{zhao2011cross,nater2011transferring}. Relation-based learning, through techniques like transferring a Markov Logic Network (MLN) or employing Relational Pathfinding (RPF), focuses on establishing effective mappings from the source model to the target domain \cite{mihalkova2008transfer,mihalkova2007mapping}. These strategies notably reduce the training time and data required to develop an accurate target domain model compared to building a model from scratch, as detailed by Mihalkova et al. \cite{mihalkova2007mapping}. This efficiency not only accelerates the learning process but also aligns with the principles of developing sustainable AI technologies by limiting the environmental footprint associated with extensive data processing and model training.

Transfer learning, particularly through parameter- and ration-based methods, holds the potential to cultivate sustainable AI systems while trading off non-discrimination and accuracy and ensuring fair transfer in the domain adaptation \cite{wang2021understanding,schumann2019transfer}. This approach enhances the interpretability of models \cite{abir2022explainable,brito2023fault}, making it easier to understand how AI makes decisions. Furthermore, explainable AI plays a crucial role in assessing both the performance and validity of models post-transfer learning, providing insights into their operation \cite{hossain2023vision}. However, challenges arise when the original, pre-trained model operates as a ``black box'' without transparency, necessitating potentially costly reprogramming efforts to enable transfer learning \cite{tsai2020transfer}. The comparative costs of this reprogramming for black-box models versus training models from scratch have yet to be thoroughly explored. 

\textbf{Machine unlearning:} Machine unlearning, as previously discussed, addresses privacy issues in machine learning by specifically eliminating individuals' data from trained models, thereby improving non-discrimination and trust in AI systems. Unlike the resource-intensive process of developing a new model from scratch to adhere to the ``right to be forgotten,'' machine unlearning methods, such as exact and approximate unlearning, offer a more eco-friendly alternative by lowering computational costs. Exact unlearning, exemplified by techniques like checkpoint training \cite{kurmanji2024towards} or SISA training with only the model trained with the subset containing to-be-forgotten data retrained \cite{bourtoule2021machine}, presents a favorable balance between unlearning accuracy and time. On the other hand, approximate unlearning employs the use of influence functions and a gradient-based deletion approach to diminish or restrict the impact of an individual's data on the model, achieving ``approximate'' statistical non-identifiability \cite{neel2021descent,izzo2021approximate}.

\textbf{Retrieval-augmented generation (RAG):} presents a cutting-edge technique in natural language processing (NLP), merging the capabilities of pre-trained language models with information retrieval methods to elevate text creation \cite{lewis2020retrieval}. This approach enables the model to dynamically access and integrate pertinent documents or data excerpts from a comprehensive corpus, infusing the text generation process with external information. The process involves using user queries to pinpoint relevant documents within a knowledge base, which are then combined with the query to form a prompt that supplies additional context for the response generation. In contrast to fine-tuning large language models (LLMs), RAG stands out as more environmentally sustainable due to its elimination of training expenses and the absence of costs associated with storing additional or modified LLMs.

Given that users can contribute to the documents retrieved, RAG has the potential to support non-discrimination through adjustments for fairness. Nonetheless, its impact on aspects such as safety, transparency, and traceability is yet to be determined.

\textbf{Model Compression:} Reducing model size without compromising performance is essential for environmentally sustainable AI, as larger models result in higher inference times and increased energy consumption and memory storage. This survey investigates four techniques applied during the training and post-training phases: pruning, quantization, low-rank factorization, and knowledge distillation. Pruning eliminates redundant parameters, neurons, layers, and filters, particularly in image processing \cite{guo2020multi,liu2021discrimination}. Quantization and Low-factor factorization can be applied to convolutional and fully connected layers, while the former reduces the number of bits for each weight and the latter decomposes a large matrix into smaller matrices of parameters with proper factorization and rank selection \cite{yamamoto2021learnable,gong2020vecq,chen2021logarithmic,hsu2022language}. Knowledge distillation, focusing on classification-based tasks, replaces large layers with smaller ones through a teacher-student learning framework \cite{ji2021show,wang2021knowledge,huang2022knowledge}.

While model compression positively impacts environmental sustainability, it adversely affects fairness measures and the model's traceability. Model compression can amplify existing biases in AI models, undermining non-discrimination efforts \cite{ramesh2023comparative,kamal2024beyond,stoychev2022effect}. Additionally, it can be used to remove watermarks within the model, thereby reducing traceability \cite{shao2024fedtracker}. The impact on transparency is multifaceted: though compression can decrease interpretability, hindering transparency \cite{joseph2020going}, explainable AI techniques can be utilized to mitigate this effect \cite{yan2024explainable,becking2020ecq,yu2023x}. Notably, compressed models can offer benefits in privacy protection \cite{chen2024privacy}.

\subsection{Traceability}
\subsubsection{Overview}

The concept of traceability pertains to the ability to relate the unique identifiable entities in a verifiable way by all parties involved, from the development phase of an AI system to its use and deployment \cite{kerrigan2022artificial}. This concept is critical to the success of AI systems in terms of ensuring transparency, accountability, and governance \cite{diaz2023connecting}.

-\textit{Regulations incorporating traceability:} One significant regulation that includes traceability as a key component is the European Union's proposed AI Act \cite{AIAct}. The AI Act emphasizes the importance of traceability for high-risk AI systems by requiring comprehensive documentation of the AI system's development, deployment, and operational phases. This documentation is critical for understanding how high-risk AI systems are developed and how they perform throughout their lifetime, enabling the verification of compliance with the Act's requirements, monitoring of operations, and post-market monitoring. The AI Act mandates that technical documentation must contain the detailed information necessary to assess the AI system's compliance with relevant requirements, including the general characteristics, capabilities, limitations of the system, algorithms, data training, testing, and validation processes used, as well as documentation on the relevant risk management system. Furthermore, the AI Act requires high-risk AI systems to technically allow for automatic recording of events (i.e., logs) throughout the system's lifetime, ensuring that activities related to the AI system can be traced back and verified.

In addition to the EU's AI Act, the U.S. Government Accountability Office (GAO) has also outlined traceability as a critical aspect of responsible AI deployment \cite{GAO}. The GAO's AI Accountability Framework includes traceability as one of its principles, advising entities to document the results of monitoring activities and any corrective actions taken to promote traceability and transparency. This approach enhances accountability by ensuring that decisions and actions taken by AI systems can be traced back to their source, facilitating oversight and the identification of issues that may require remediation.

These regulations and frameworks recognize the significance of traceability in maintaining the transparency, accountability, and governance of AI systems, particularly those that pose significant risks to individuals and society.

-\textit{The role of traceability in AI regulation:}
The concept of traceability is crucial in AI regulation, providing a fundamental structure that guarantees AI systems are effectively monitored and scrutinized, held accountable, and conform to ethical norms and legal mandates \cite{kroll2021outlining}. By enabling the ability to track and understand all actions, decisions, and processes involved in the AI lifecycle, traceability fosters a higher degree of transparency. This is essential not only for the creators of AI systems but also for regulators and the public, ensuring that any decisions made by AI can be scrutinized and understood, thereby upholding accountability, especially when these decisions significantly impact individuals or society at large.

In the realm of AI regulations and guidelines, traceability is often a stipulated requirement. This entails maintaining comprehensive records of the data, models, and processes utilized throughout the development and deployment phases of AI systems. Such meticulous documentation simplifies the process of demonstrating \textit{compliance with both legal and ethical standards} \cite{kroll2021outlining}. For instance, the EU AI Act emphasizes the importance of keeping detailed records as a means to facilitate adherence to regulatory frameworks, thereby underscoring the critical role of traceability in regulatory compliance \cite{AIAct}.

Furthermore, traceability is integral to fostering \textit{ethical and responsible AI development and use} \cite{peters2020responsible}. It provides the necessary infrastructure to audit and review the behavior of AI systems thoroughly. This capability is crucial for identifying, analyzing, and rectifying biases, errors, or any unintended consequences that may emerge throughout an AI system's lifecycle. Consequently, traceability supports the operationalization of ethical principles and the practice of responsible AI, ensuring that AI systems are developed and utilized in a manner that aligns with societal values and norms.

From a \textit{risk management perspective}, traceability equips organizations with the tools to identify and mitigate potential issues early on \cite{steimers2022sources}. It allows for the continuous monitoring and evaluation of AI systems to ensure their operation remains within the intended ethical and regulatory parameters. By enabling early detection and correction of problems, traceability significantly contributes to minimizing risks associated with AI systems, thereby ensuring their safe and effective deployment.

Lastly, traceability plays a crucial role in \textit{enhancing trust} in AI technologies among users and the broader public. The knowledge that the workings and decisions of an AI system can be traced back to their origins fosters confidence in the technology's reliability and fairness \cite{bedue2022can}. In an era where trust in AI is paramount for widespread adoption, traceability emerges as a key factor in building and maintaining this trust, thereby facilitating a more receptive environment for AI technologies.
\begin{table*}[!htb]
\caption{Mechanism/techniques influencing AI traceability and the discussion on environmental sustainability versus safety, transparency, non-discrimination and environmental sustainability (``-'': negative impact, ``+'': positive impact)}
\label{tab:Trace_impact}
\centering
\scriptsize
\begin{tabular}{p{4.5cm}p{1.3cm}p{2cm}p{3.2cm}p{2.5cm}}
\toprule
\textbf{Papers} & \textbf{Category} & \textbf{Techniques} & \textbf{Impact} & \textbf{Discussion with the Rest Aspects in AI Regulation} \\
\midrule \midrule

\begin{tabular}[c]{@{}p{5cm}@{}}\cite{kavasidis2023federated,chen2023privacy,li2020review}\end{tabular} & 
\parbox{1.3cm}{Blockchain} & 
Blockchain-based Federated Learning & 
\parbox{3.2cm}{Recording each node's training data and process for backtracking and audit} & 
\parbox{2.5cm}{Environmental sustainability (+), safety (+), transparency (+)} \\[5pt]
\midrule

\cite{barni2001watermark,sahu2024multimedia} & 
\multirow{6}{*}{\parbox{1.3cm}{Water marking}} & 
Direct embedding in data & 
Enhancing source tracking and audit trails & 
\multirow{6}{*}{\parbox{2.5cm}{Safety (+), transparency (+)}} \\[5pt]
\cmidrule(r){1-1} \cmidrule(l){3-4}

\cite{min2024watermark,nie2024deep} & & 
Model watermarking & 
Providing IP protection \parbox{3.5cm}{and regulatory compliance} & \\[5pt]
\cmidrule(r){1-1} \cmidrule(l){3-4}

\begin{tabular}[c]{@{}p{4.5cm}@{}}\cite{zhang2024dual,abdelnabi2021adversarial}\end{tabular} & & 
Adversarial watermarking & 
\parbox{3.2cm}{Proving ownership when AI models are subjected to various attacks} & \\[5pt]
\midrule

\cite{mora2021traceability} & 
\parbox{1.3cm}{AI systems reproduceability} & 
Practices and tool support & 
\parbox{3.2cm}{Using relevant tools, practices, and data models for traceability in their connection to building AI models and systems} & 
\parbox{2.5cm}{Safety (+), transparency (+), non-discrimination (+)} \\[5pt]
\midrule

\begin{tabular}[c]{@{}p{4.5cm}@{}}\cite{lin2006open,cui2024rethinking}\end{tabular} & 
\parbox{1.3cm}{Open-source licenses} & 
Regulation & 
\parbox{3.2cm}{Controlling the misuse and unauthorized use of open-source code} & 
Transparency (+) \\[5pt]
\bottomrule
\end{tabular}
\end{table*}

\subsubsection{Traceability versus safety, transparency, non-discrimination, and environmental sustainability}

Table~\ref{tab:Trace_impact} summarizes the approaches used to prompt AI traceability with its impacts on the other four aspects of ethical AI.

\textbf{Blockchain:} Blockchain technology employs its consensus mechanism and cryptographic technologies to enable users to record data on an immutable ledger, thus ensuring the integrity and traceability of the data. In the era of artificial intelligence, blockchain methods are increasingly used to store and share training data, model parameters, and training processes, enhancing traceability, privacy, accountability, and audibility \cite{kavasidis2023federated}.

Functioning as a distributed ledger maintained by a peer-to-peer network, blockchain is often integrated into a federated learning (FL) framework. This approach decentralizes the model training process, distributing it across various nodes to maintain data privacy and security \cite{li2020blockchain,kavasidis2023federated,chen2023privacy,li2020review}. In regulated industries such as healthcare and pharmaceuticals, where process reproducibility is crucial --- including model reconstruction --- blockchain provides a robust solution to meet these stringent requirements \cite{kavasidis2023federated,li2020review}. More specifically, they deploy a multi-blockchain-based platform to create a comprehensive audit trail for all activities associated with the FL model training process \cite{kavasidis2023federated}. In this setup, each node in the model uses a blockchain to store the training data and intermediate data derived from training sessions, each secured with hash values. Concurrently, a global blockchain, managed by a smart contract, orchestrates the overall training process. To minimize the storage burden on the blockchain, model parameters are kept in an InterPlanetary File System (IPFS) distributed network, with the ledger recording the corresponding IPFS-created addresses \cite{chen2023privacy}. This approach significantly reduces the environmental impact of heavy data sharing and transmission in blockchain operations. Simultaneously, distributing the processing burden of the FL framework across multiple computational nodes minimizes both operational and computational overhead. 

Since blockchain techniques ensure not only traceability but also privacy on an immutable ledger, it is crucial for regulated companies like aerospace and pharmaceutical companies. These sectors require maintaining the highest quality standards for their products and the ability to reconstruct the entire production chain to gain crucial insights into the causes of any failures \cite{kavasidis2023federated}. Therefore, it improves the model's safety and transparency.

\textbf{Watermarking:} In the context of AI that involves embedding a unique, invisible marker or pattern into AI-generated outputs (such as images, videos, or text) or within the AI model itself. This marker is designed to be robust against modifications and should be retrievable even after the data undergoes transformations or compression \cite{regazzoni2021protecting}.

Watermarks are subtly incorporated directly into content, such as images or video frames, through a process known as direct embedding \cite{barni2001watermark,sahu2024multimedia}. This method ensures the watermarks are both imperceptible to users and recoverable when needed. It is a technique frequently utilized in various forms of media and official documents. Model watermarking, on the other hand, involves adjusting a neural network’s internal parameters-like specific weights or biases-to embed a watermark \cite{min2024watermark}. This is done in such a way that the model’s performance remains unaffected \cite{nie2024deep}. The presence of the watermark can be confirmed through specific queries that prompt the model to produce pre-defined responses, revealing the watermark. Adversarial watermarking represents a more advanced technique \cite{zhang2024dual,abdelnabi2021adversarial}. In this method, the process of embedding a watermark involves the creation of adversarial example inputs specifically designed to challenge the model. These inputs are subtly modified to include the watermark and are used during the model’s training phase. The model is thus trained to recognize these examples and respond in a specific manner, embedding the watermark more thoroughly within its operational framework. This type of watermarking embeds it deeply within the model’s decision-making patterns, making it particularly robust.

 Watermarking can enhance the safety of AI systems by ensuring that the models and their outputs have not been tampered with. This is particularly critical in applications like autonomous vehicles or healthcare, where data integrity is crucial for making safe decisions. Watermarking can make the origins of AI-generated content clearer, which is essential for transparency. Users and regulators can verify where an AI output came from and whether it has been altered from its original form.

\textbf{Reproducibility:} The reproducibility of AI systems facilitates traceability by ensuring that the processes and results of AI models can be reliably duplicated under similar or different conditions. This capability is vital for verifying and validating the methodologies and outcomes involved in AI research and applications, thereby enhancing traceability, transparency and trustworthiness.

Several tools have been developed to achieve reproducibility, which, in turn, supports traceability \cite{mora2021traceability}. ModelDB \cite{vartak2016modeldb} is an open-source system designed for versioning machine learning models, allowing users to index, track, and store modeling artifacts for later reproduction, sharing, and analysis. It emphasizes experiment tracking and provides a web-based interface for the visual representation and analysis of models.
Code Ocean \footnote{https://codeocean.com/}, Whole Tale \footnote{https://wholetale.org/}, and The Renku Project \footnote{https://datascience.ch/renku/} are among the numerous online tools aimed at enhancing reproducibility. These platforms leverage cloud storage and containerization technologies, such as Docker, to capture the research environment fully. This enables the reuse, sharing, and reproduction of the entire research process. Code Ocean merges leading tools, languages, and environments to offer an end-to-end workflow focused on reproducibility. Whole Tale is a free, open-source platform that captures data, code, and the complete software environment, aspiring to redefine how computational and data-driven science is conducted and reproduced. The Renku Project combines a web platform with a command-line interface to support reproducibility, reusability, and collaboration.
A broader set of tools supporting ``methods reproducibility'' research includes ZenML \footnote{https://zenml.io/}, Binder \footnote{https://mybinder.org/}, DVC (i.e., Data Version Control) \footnote{https://dvc.org/}, Taverna \footnote{https://taverna.incubator.apache.org/}, Kepler \footnote{https://kepler-project.org/}, and VisTrails \footnote{https://www.vistrails.org/}. These tools are designed to manage different aspects of the computational research lifecycle, including environment setup, code execution, data management, and the tracking of provenance and metadata. These tools provide essential capabilities for ensuring that AI research and applications are reproducible, which in turn supports the traceability of AI systems by documenting and validating the steps, data, and outcomes involved in model development and deployment.

Traceability in AI systems involves maintaining comprehensive records of the data, decisions, processes, and methodologies used throughout the AI lifecycle. This documentation is crucial for ensuring AI safety. It allows developers and stakeholders to understand how an AI system was built, trained, and deployed, making it easier to identify, diagnose, and rectify any safety issues that may arise. Traceability ensures that AI systems can be audited and reviewed for safety compliance, potentially preventing harm to users and the public. Transparency in AI refers to the openness and clarity regarding how AI systems operate, make decisions, and are developed. Traceability supports transparency by providing a detailed record of the AI development process, including the data used for training, algorithms applied, and decision-making processes. This information is essential for stakeholders to assess the reliability and trustworthiness of AI systems. Transparency, supported by traceability, fosters trust among users, regulators, and the public by making AI operations understandable and open to scrutiny. Traceability supports fairness by documenting the data, algorithms, and decision-making processes used, allowing for the examination and correction of potential biases. By maintaining transparent records, stakeholders can audit AI systems to identify and mitigate unfair practices, ensuring that AI technologies produce equitable outcomes for all users.

\textbf{Open-source license:} An open-source license is a type of license for computer software and other products that allows the source code to be used, modified, and distributed by anyone. Open-source licenses are designed to encourage collaboration and sharing, promoting the development of the software in a community-driven manner. These licenses allow the software to be freely used, modified, and shared under defined terms and conditions \cite{steiniger20132012}.

There are various types of open-source licenses, each with its own specific terms that define how the software can be used, modified, and distributed. Some of the most common open-source licenses include (1) MIT License \footnote{https://opensource.org/license/mit}: One of the most permissive and straightforward open-source licenses, allowing almost unrestricted freedom to use, modify, and distribute the software, provided that the license and copyright notice are included with any substantial portions of the software; (2) Apache License \footnote{https://www.apache.org/licenses/LICENSE-2.0}: Allows the user to freely use, modify, and distribute the software, with the condition that any modifications are documented, and the original copyright and license notices are provided with any distributions. It also grants a patent license to contributors; (3) Source Distribution (BSD)  License \footnote{https://opensource.org/license/bsd-3-clause}: The BSD License is another permissive open source license that maintains license notices and copyrights while allowing larger or licensed works to be distributed without source code under different license terms. 

Open-source licenses require that the source code be made available to the public. This transparency allows researchers, developers, and users to examine the algorithms, data processing methods, and decision-making processes within AI systems \cite{mckay2022ai}. It helps in understanding how these systems work, identifying potential biases, errors, or vulnerabilities, and ensuring that the AI behaves as intended. By allowing anyone to access, modify, and distribute the source code, open-source licenses foster a collaborative environment. Open-source AI projects also enable other researchers to reproduce and verify the results claimed by the original developers. This is a fundamental aspect of scientific research that ensures the reliability and validity of AI technologies. Lastly, open-source projects typically maintain extensive documentation and use version control systems. This practice enhances traceability by providing a detailed history of changes, updates, and modifications made to the AI system over time. 

\subsection{Non-discrimination}
\subsubsection{Overview}

\begin{table*}[!htbp]
\caption{Mechanism/approaches influencing AI non-discrimination and the discussion on non-discrimination versus safety, transparency, environmental sustainability, and traceability (``-'': negative impact, ``+'': positive impact)}
\label{tab:nondis_impacts}
\centering
\scriptsize
\begin{tabular}{p{0.5cm}p{1cm}p{5cm}p{5cm}p{2.5cm}}
\toprule
\textbf{Stage} & \textbf{Category} & \textbf{Papers} & \textbf{Explanation} & \textbf{Trade-off} \\ 
\midrule \midrule

\multirow{11}{*}{Pre} & 
\multirow{3}{*}{Sampling} & 
\cite{feldman2015certifying,adler2018auditing,dwork2018decoupled} & 
Implements statistical parity through balanced data sampling and removes disparate impact by modifying the distribution of protected attributes in training data. Key methods include stratified sampling and balanced mini-batch creation. & 
Safety (+), transparency (+) \\ 
\cmidrule(lr){2-5}

& \multirow{4}{*}{Relabeling} & 
\cite{luong2011k,miron2020addressing,wang2019repairing} & 
Corrects biased labels in training data through KNN-based label propagation and gradient descent optimization. Focuses on identifying and rectifying mislabeled instances that contribute to discriminatory outcomes. & 
Transparency (+), environmental-sustainability (+) \\
\cmidrule(lr){2-5}

& \multirow{4}{*}{\begin{tabular}[c]{@{}p{1cm}@{}}Represen-tation\end{tabular}} & 
\cite{Bower2017FairP,brunet2019understanding,kairouz2019generating,du2018data} & 
Transforms raw data into fair representations through dimensionality reduction and feature engineering. Employs convex optimization to learn representations that maximize task performance while minimizing correlation with protected attributes. & 
Transparency (+), traceability (+), safety (+) \\
\midrule

\multirow{13}{*}{In} & 
\multirow{5}{*}{\begin{tabular}[c]{@{}p{1cm}@{}}Treatment-driven\end{tabular}} & 
\cite{zafar2017fairness,zafar2017fairnessbeyond,bellamy2018ai,cheng2021fairfil,zhou2021contrastive} & 
Enforces equal treatment by modifying model architecture and loss functions. Incorporates fairness constraints directly into optimization objectives through adversarial training and regularization terms that penalize discriminatory decisions. & 
Safety (+), environment-sustainability (-), transparency(+), traceability (+) \\
\cmidrule(lr){2-5}

& \multirow{4}{*}{\begin{tabular}[c]{@{}p{1cm}@{}}Impact-driven\end{tabular}} & 
\cite{zhang2018mitigating,wadsworth2018achieving,edwards2015censoring,elazar2018adversarial,sweeney2020reducing} & 
Focuses on equalizing model outcomes across protected groups through adversarial debiasing and demographic parity constraints. Explicitly optimizes for balanced prediction distributions regardless of sensitive attributes. & 
Safety (+), environment-sustainability (-) \\
\cmidrule(lr){2-5}

& \multirow{4}{*}{Hybrid} & 
\cite{madras2018learning,kim2021counterfactual,park2021learning} & 
Combines treatment and impact-driven approaches through multi-objective optimization. Uses disentangled representations and counterfactual fairness to simultaneously address disparate treatment and impact. & 
Transparency (+), safety (+) \\
\midrule

\multirow{12}{*}{Post} & 
\multirow{4}{*}{\begin{tabular}[c]{@{}p{1cm}@{}}Input correction\end{tabular}} & 
\cite{agarwal2019fair,kiritchenko2018examining} & 
Adjusts test inputs through targeted perturbation and feature transformation to ensure fair predictions. Implements systematic modifications to input data while preserving task-relevant information. & 
Environmental-sustainability (-), transparency (+) \\
\cmidrule(lr){2-5}

& \multirow{4}{*}{\begin{tabular}[c]{@{}p{1cm}@{}}Classifier correction\end{tabular}} & 
\cite{adler2018auditing,mcnamara2017provably,anders2020fairwashing} & 
Modifies trained model's decision boundaries through post-hoc calibration and threshold adjustment. Implements fairness constraints while maintaining model performance through minimal architectural changes. & 
Environmental-sustainability (-), safety (+) \\
\cmidrule(lr){2-5}

& \multirow{4}{*}{\begin{tabular}[c]{@{}p{1cm}@{}}Output correction\end{tabular}} & 
\cite{mehrabi2021attributing,jang2022group} & 
Applies post-processing to model predictions using group-specific thresholds and rejection sampling. Ensures statistical parity in final outputs while maintaining prediction quality. & 
Environmental-sustainability (-), transparency (+) \\
\bottomrule
\end{tabular}
\end{table*}

Non-discrimination in the context of the EU AI Act means developing and using AI systems that promote diversity, equal access, and gender equality while avoiding discriminatory impacts and unfair biases that are prohibited by Union or national law. 

-\textit{Non-discrimination in the AI Act:}
In the EU AI Act, the concept of non-discrimination, also refer as fairness, is not explicitly defined in a single section, but rather, it permeates various aspects of the regulation through provisions aimed at ensuring that AI systems do not create or perpetuate discrimination or bias. Firstly, the Act emphasizes the prevention of discrimination by requiring that high-risk AI systems undergo rigorous assessment processes to ensure they do not produce biased outcomes \cite{AIAct}. This includes testing, validation, and documentation to demonstrate that these systems can handle data fairly without leading to discriminatory results.
Secondly, non-discrimination is also promoted through transparency and explainability requirements, particularly for high-risk AI systems. The Act mandates that operators provide clear information on AI systems' functioning, capabilities, and decision-making processes \cite{AIAct_new}. This transparency helps stakeholders understand how decisions are made, thereby promoting non-discrimination in the operation of AI systems. In addition, proper management of data used by AI systems is critical to ensuring non-discrimination. The Act requires high-quality data governance practices to prevent biases arising from data misuse or poor quality. This includes measures for the accuracy, reliability, and representativeness of the data sets used, which directly influence the non-discrimination of the AI system's outcomes. Lastly, the regulation explicitly prohibits AI practices that could manipulate or exploit vulnerable groups or otherwise lead to unfair outcomes. For example, AI systems designed to exploit vulnerabilities of individuals based on age, economic situation, or disabilities are forbidden. By integrating these principles, the AI Act seeks to ensure that AI systems used within the EU contribute to equitable outcomes, do not reinforce unfair biases, and respect the principle of non-discrimination as laid out in the broader framework of EU law.

-\textit{Addressing bias throughout the lifecycle of an AI system:} As shown in Table \ref{non-discrimination}, we organize strategies for reducing bias in AI systems into three distinct stages, namely pre-processing, in-processing, and post-processing, each focusing on a different stage of the development and deployment process for AI systems. Considerable mitigation at the pre-processing stage involves carefully examining and preparing the data. Related methods ensure the data is representative of all demographics and does not contain historical biases or skewed distributions that could influence the AI's decision-making process.
At the in-processing stage, 
non-discrimination constraints are integrated directly into the algorithm's learning process. By adjusting the learning algorithms to account for equity, in-processing aims to prevent the AI system from perpetuating existing biases that might be present in the training data. 
The final stage of bias mitigation focuses on the outputs of AI systems. After an AI model has been trained, post-processing techniques are applied to adjust its decisions to ensure they adhere to non-discrimination principles.

-\textit{Non-discrimination-driven approaches to public safety:} Safety emphasizes the prevention of harm to individuals or data breaches due to technical failures or design flaws in AI systems, including avoiding unfair risks to specific groups due to bias. The interaction between non-discrimination and safety ensures that AI systems are safe and fair for all users. Below are particular algorithms and research examples illustrating this tension. For instance, striving for demographic parity through equal male and female parole rates may inadvertently disadvantage lower-risk female inmates to fulfill this ratio, thereby breaking the non-discrimination principle of equalized odds \cite{berk2021fairness}. The application of machine learning in critical sectors such as the judiciary, welfare systems, and autonomous driving underscores the myriad ways in which AI systems, imbued with inherent biases, can impact daily life and the subtleties of biases in AI and robotics infiltrating real-world scenarios \cite{howard2018ugly}, emphasizing the non-discrimination imperative for researchers and engineers to anticipate downstream effects in AI system development. Acknowledging the predictive value of gender in such contexts could lead to a decision-making paradigm that is unfair to minority groups, without secured public safety outcomes. By implementing these approaches, AI systems can contribute to public safety through equitable law enforcement, unbiased emergency response, inclusive public health measures, and fair access to public services. Ultimately, non-discriminatory AI helps reduce social tensions, improve trust in institutions, and ensure that technological advancements benefit all members of society equally \cite{mitchell2021algorithmic}.

\subsubsection{Non-discrimination versus transparency, traceability, safety, and environmental sustainability} \label{subsubsec:nondis_impacts}
In the framework of the European Union's AI Act, the principle of non-discrimination, along with traceability, transparency, safety, and environmental sustainability, constructs a comprehensive framework to ensure the ethical and societal responsibility of AI systems. This section elucidates how these intertwined principles foster a fair, secure, and sustainable deployment of AI technologies.

\textbf{Enhancing non-discrimination through increased transparency:} Transparency is closely linked to the principle of non-discrimination, as it enhances the visibility of system operations, allowing for external assessments of bias or unfair practices. Beyond aiding stakeholders in identifying potential biases, transparency also bolsters trust in the non-discrimination and reliability of AI systems.

Transparency and explanation \cite{grgic2018human}\cite{srivastava2019mathematical} are fundamental in fostering trust and understanding in machine learning non-discrimination. 
By identifying and addressing the underlying causes of bias, causal methods can help reveal underlying biases and improve decision-making transparency. While Inverse Propensity Scoring (IPS) techniques \cite{bonner2018causal} offer a useful approach for enhancing fairness in AI systems by tackling selection bias and promoting more equitable datasets, they are not without drawbacks. These methods, though practical, struggle to adapt to changes in observational patterns and are frequently designed for particular use cases rather than being universally applicable.
By introducing small changes to input data, perturbation methods can reveal hidden biases in AI models, helping identify areas where discrimination may occur. Studies using perturbation techniques have demonstrated that implementing measures to ensure non-discrimination does not substantially reduce accuracy in AI systems \cite{patro2020incremental}\cite{ekstrand2018exploring}. This finding supports the use of such interventions to enhance privacy protection, transparency and fairness in AI applications.
Interpretable models \cite{liu2022trustworthy} further non-discrimination by making decision processes transparent, advocating for a cohesive approach to creating fair and transparent algorithmic systems \cite{li2006t}.

\textbf{Enhancing non-discrimination through increased traceability:} At the heart of traceability lies the commitment to ensuring that the decision-making pathways, data sources, and algorithmic logic of AI systems are transparently documented and traceable. In essence, traceability serves as an implementation mechanism for non-discrimination, supporting the assessment and verification of non-discrimination through detailed documentation of decision processes. 

Studies \cite{brunet2019understanding} introduce a pre-processing strategy to mitigate bias by altering or removing bias-originating documents during training, addressing word embedding bias. Concurrently, research \cite{kilbertus2018blind} develops an encryption method for sensitive user data, enhancing security and privacy while allowing non-discrimination verification without exposing data to unauthorized use or access.

\textbf{The impact of non-discrimination AI approaches to sustainable AI:} Repairing or rebuilding a fair model is typically time-consuming, particularly when post-processing is required \cite{jang2022group,adler2018auditing,agarwal2019fair}. Adversarial learning techniques, such as adversarial debiasing, eliminate discrimination using a discriminator, but introducing additional model training and inference processes is not environmentally friendly \cite{zhang2018mitigating,beutel2017data,sweeney2020reducing}. However, some repair methods are relatively more sustainable than rebuilding. For example, using a suitable algorithm to learn counterfactual distributions can repair a black-box classifier without the need for retraining, offering an environmentally sustainable way to restore the model's fairness \cite{wang2019repairing}.
\section{Discussion and future directions}


Addressing the open problems of ethical AI requires concerted efforts across multiple stakeholders, 
including governments, private sectors, academia, and civil society. 
By fostering an environment that prioritizes ethical considerations in AI development and use, 
it is possible to harness the benefits of AI technologies while mitigating their risks and ensuring they contribute positively and ethically to society. 
In the following, 
we first summarize the open problems associated with the AI Act from the perspective of technical efforts to promote human-centric principles, including safety, traceability, transparency, environmental sustainability, and non-discrimination. 
We then explore potential approaches for regulating AI systems in the future.

\subsection{AI Act's technical challenges for human-centric design} \label{subsec:discuss_challenges}
\textbf{Strengthening AI systems against supply-chain vulnerabilities:} Addressing supply-chain vulnerabilities in AI systems extends beyond protecting against direct attacks to include securing the entire ecosystem surrounding AI development, from the data sources and software libraries to the deployment environments. 
These vulnerabilities may arise from compromised data leading to poisoned models, 
tampered software libraries inducing backdoor threats, 
or adversarial manipulations undermining model integrity, availability, and privacy. 
Effective mitigation requires a comprehensive approach that combines advanced technical strategies, 
like adversarial training for resilience against attacks \cite{chen2022adversarial}, 
\cite{abdelnabi2021adversarial}, 
and rigorous data sanitization \cite{venkatesan2021poisoning} to prevent malicious data injection, 
with robust procedural and policy frameworks. 
These measures must account for the inherent trade-offs between model accuracy and robustness, 
and the dynamic nature of AI threats necessitates continuous vigilance and adaptation. 
Enhancing the security of AI systems against supply-chain vulnerabilities thus involves a concerted effort across the AI community to foster secure, reliable, and trustworthy AI through ongoing collaboration, innovation, and a strong commitment to security best practices.

\textbf{The dual-edged nature of technologies in ethical AI promotion:} Promoting ethical AI underscores the nuanced balance between harnessing innovative technological advances to foster ethical AI practices and navigating the inherent challenges these technologies may pose. 
On one side, technologies like machine learning algorithms, blockchain, and data analytics tools can significantly enhance transparency, accountability, and fairness in AI systems, laying the foundation for more ethical AI development. 
They provide mechanisms for unbiased decision-making, environmental friendliness, secure data sharing, and enhanced privacy, which are crucial for ethical standards. 
For example, transfer learning, as discussed in Section~\ref{subsubsec:ef_impacts}, 
typically has a positive influence on the development of AI models, particularly in terms of environmental sustainability, non-discrimination, and transparency. 
Nonetheless, not all transfer learning approaches yield beneficial outcomes for environmentally sustainable development, and some are necessary only in specific scenarios. 
For example, instance-based transfer learning often has a detrimental environmental impact, while feature-based transfer learning might have a slightly negative effect. 
In situations where the training set is small, and label information originates solely from the source domain, users might be constrained to opt for either instance-based or feature-based transfer learning over more eco-friendly alternatives. 
According to \cite{niu2020decade}, other environmental sustainability transfer learning methods can be chosen when the label information of the target-domain instances is available.
Otherwise, a trade-off between environmental sustainability and performance shall be considered \cite{wu2022cost}. 
Additionally, the efficiency of instance-based or feature-based transfer learning in enhancing non-discrimination and transparency while remaining cost-effective is yet to be determined.

Another example is adversarial training, as discussed in Section~\ref{subsubsec:safety_impacts} and Section~\ref{subsubsec:nondis_impacts}. 
It significantly enhances the safety and non-discrimination aspects of AI model development, respectively \cite{madry2017towards,tsipras2018robustness,zafar2017fairness,beutel2017data}. 
This method strengthens the model's resilience or reduces its dependency on certain specific features by introducing adversarial examples during the training phase. 
However, while aimed at promoting model fairness, adversarial training might inadvertently compromise the model's performance in critical scenarios, which leads to potential threats in safety \cite{beutel2017data}, such as identifying illegal weapon possession \cite{zafar2017fairness}. 
Adversarial training with crafted adversarial examples introduces the features that the model learns to treat independently, enhancing the system's transparency and traceability by offering clear insights into the model's robustness/defenses against particular adversarial inputs \cite{madry2017towards,tsipras2018robustness}. 
Further, adversarial training may introduce a more complex training process, resulting in more computational cost and potentially conflicting with environmental sustainability objectives \cite{zafar2017fairness}.

Differential privacy (DP) is the third technique discussed due to its dual-edged nature. 
Generally, DP enhances AI transparency and safety but poses challenges for developing environmentally friendly AI and complicates traceability, with a complicated effect on fairness.
This technique is adaptable and applicable across various stages of AI development, including training data preprocessing, feature engineering, model training, and prediction, primarily to ensure privacy-preserving data analysis \cite{dwork2006differential,zhu2020more}. 
Such applications help AI providers establish suitable levels of transparency tailored for different users or customers. 
In addition, with DP techniques, the AI model can be more robust and safe by mitigating various attacks like membership inference attacks and model inversion attacks \cite{truex2019effects,salim2022perturbation}. 
However, incorporating DP increases the complexity of models, learning tasks, and datasets, particularly during the model training phase, which can contradict the principles of environmental sustainability \cite{truex2019effects}. As for traceability, although the DP-based model is more complex to understand than the typical model, the DP-based explainable model shows more private interpretability. Patel et al.~\cite{patel2022model} proposed an optimization method to minimize the total privacy loss while maintaining a high explanation quality. 
The impact of DP on the fairness of AI systems is complex. 
On the one hand, it supports fairness by enabling re-sampling of training data \cite{zhu2020more}, which helps maintain balance across different groups. 
On the other hand, using differentially private training data can inadvertently introduce biases, potentially affecting fairness negatively \cite{tran2021decision}, though these effects might be mitigated by perturbing the model’s outputs \cite{mangold2023differential}. 
Thus, while DP contributes positively to transparency and safety, its influence on environmental sustainability, traceability, and fairness in AI necessitates careful consideration and balanced application.

\textbf{Ensuring Ethical Development in Large Language Models:}
The importance of data in the development of Large Language Models (LLMs) cannot be overstated. 
The correlation between the growth of these models and the exponential increase in required training data is evident in their evolution. 
Although detailed disclosure of data sources by LLMs remains rare, the instances where such transparency exists reveal the vast data consumption involved in training \cite{touvron2023llama}. 
The emergence of chatbots \cite{chatgptopenai} marks a significant technological leap, offering potential across a wide range of business applications, from entertainment to critical sectors. 
However, this advancement comes with challenges, notably the vulnerability of LLMs to prompt injections \cite{greshake2023not}, a technique that can elicit undesired responses. 
Recent study has identified inherent limitations in strictly censoring LLMs, highlighting the need for alternative risk management strategies such as implementing controlled access points for models and enhancing cybersecurity measures\cite{glukhov2023llm}.
Despite the impressive capabilities demonstrated by chatbots in specific tasks, the technology is still in its infancy and requires cautious deployment, especially in scenarios where trustworthiness is paramount. The need for continuous monitoring underscores the technology's emerging status and the inherent challenges in ensuring ethical AI development. 
Future directions should focus on increasing transparency around data sources, enhancing security measures to counter vulnerabilities, and adhering to rigorous ethical standards to foster trust and reliability in LLM applications.

\textbf{Trade-offs between the attributes of ethical and trustworthy AI:} To navigate towards the goal of ethical AI, it is paramount to recognize and address the intricate balance among the diverse attributes that underpin the trustworthiness of AI systems. 
These attributes include human-centric principles that are defined in regulation as well as some that extend beyond, such as accuracy, susceptibility to adversarial attacks, explainability, privacy, and robustness to adversarial attacks. 
A singular focus on optimizing one attribute, such as accuracy, often results in compromises in other critical areas like fairness and vulnerability to attacks \cite{jagielski2019differentially}. 
This delicate balancing act between various desirable qualities highlights a key challenge in AI development: the inherent trade-offs that must be managed. 
For instance, enhancing an AI system's adversarial robustness might inadvertently lower its accuracy and fairness \cite{wang2020once}.

The path forward involves a multifaceted approach that does not seek to maximize performance in one attribute at the expense of others but rather aims for a harmonious balance that upholds the principles of AI regulation. 
This requires a comprehensive understanding of the interplay between different AI system characteristics, necessitating ongoing research into characterizing and navigating the trade-offs between them. 
As AI technology becomes increasingly integrated into various facets of modern life, the importance of such research grows, underscoring the need for innovative solutions that ensure AI systems are trustworthy by being fair, secure, transparent, traceable, and environmentally friendly. 
Future directions in achieving ethical AI must focus on developing methodologies and technologies that enable this balance, ensuring AI systems are designed and deployed in a manner that earns and maintains public trust \cite{ai2023artificial}.

\subsection{Future AI regulation directions}
\textbf{Rapid AI technological development vs slow and robust regulatory frameworks:} 
With the widespread application of AI, a comprehensive regulatory framework is essential to govern AI development and deployment, encompassing various laws related to the environment and labor relations. 
This survey focuses on ethical issues such as safety, transparency, traceability, non-discrimination, and environmental sustainability. 
However, the rapid advancement of AI results in continuous breakthroughs and widespread deployment across multiple domains, creating a regulatory lag and involving a complex array of stakeholders, as shown in Figure ~\ref{fig:human_centric_ai}.
These stakeholders include enterprises, researchers and developers, regulators and policymakers, end-users, and customers. 
AI technologies enable rapid innovation cycles with frequent updates and iterations to enhance performance and capabilities, which can sometimes surpass researchers' and developers' ability to fully understand and address potential ethical concerns \cite{blauth2022artificial,nasr2023scalable}. 
For example, jailbreaking prompts have recently emerged as an effective method that may missed by end-users and customers to bypass security restrictions and generate harmful content that was originally intended to be prohibited \cite{yu2024don}.

\textbf{Required technical complexity and global standardization:}
Robust regulations and legislative acts necessitate a consensus on values and moral perspectives, as well as a deep understanding of technical complexities to achieve AI goals. 
This complexity stems from the need to tackle a diverse array of issues while ensuring that AI technologies are safe, transparent, traceable, non-discriminatory, and environmentally sustainable.
As discussed in the previous section ~\ref{subsec:discuss_challenges}, it is crucial to identify technical challenges and trade-off solutions to comprehensively address ethical AI goals. 
Additionally, given the rapid misuse and abuse of LLMs \cite{blauth2022artificial}, it is essential to understand related technologies to identify associated risks and corresponding solutions. 
Our survey provides a comprehensive analysis of AI technologies regarding various ethical issues.

Furthermore, AI systems are currently developed and implemented under varying national regulations as described in Section \ref{introduction}, which leads to fragmentation. 
The lack of unified standards across countries creates disparities in legal certainty and market access for AI operators. Furthermore, the Act highlights the need for international cooperation to achieve consistent standards and regulatory practices. 
Efforts include pursuing international agreements and mutual recognition of conformity assessments to harmonize AI regulations globally.

\textbf{Interdisciplinary collaboration beyond computer science:} An in-depth understanding of AI trustworthiness necessitates not only the development of newer and better AI technologies but also a comprehensive grasp of how these technologies interact with human society. 
This multifaceted approach calls for collaboration across various disciplines far beyond computer science. 
AI practitioners should collaborate closely with domain experts whenever AI technologies are deployed in real-world scenarios that impact people, such as in medicine, finance, transportation, and agriculture. 
Domain experts can provide crucial insights into industry-specific challenges and help ensure that AI applications are both effective and responsible. 
Furthermore, AI practitioners should seek advice from social scientists better to understand the often unintended societal impacts of AI. 
Social scientists can help identify and address issues such as the effects of AI-automated decisions, job displacement in AI-impacted sectors, and the influence of AI systems on social networks. 
By working together, AI developers and social scientists can develop strategies to mitigate negative outcomes and enhance the positive impacts of AI. 
Lastly, it is essential for AI practitioners to carefully consider how AI technology is presented to the public and interdisciplinary collaborators. 
Transparent and honest communication about the known limitations of AI systems is crucial for building trust and ensuring responsible use. 

\textbf{Societal impacts:} The regulation of AI is essential for balancing technological advancement with societal welfare, particularly regarding job displacement. 
Current AI regulations face several limitations in addressing societal impacts effectively. 
Many regulations are reactive rather than proactive, often responding to issues only after they have arisen. 
This delayed action can leave displaced workers without immediate support or retraining opportunities, exacerbating the negative effects of job loss. 
Additionally, the lack of comprehensive frameworks means that regulations tend to focus on specific sectors or technologies without considering the broader economic and social implications, leading to gaps in protection for workers across different industries.

\section{Conclusion}

In the survey of the AI Act and its technical and regulatory dimensions, we have journeyed through a landscape where the pillars of AI-safety, transparency, non-discrimination, traceability, and environmental sustainability-are both the foundation and the horizon of regulatory endeavours. 
Our exploration has not only mapped out the current state of AI advancements and their interplay with regulatory frameworks but also illuminated the intricate pathways through which these principles can be harmonized to foster AI systems that are not only innovative but also ethical, equitable, and sustainable.

As we conclude this review, it becomes evident that the AI Act is more than a legislative framework; it is a compass guiding the AI community through the complexities of ethical AI creation and utilization. 
The Act's focus on safety, transparency, non-discrimination, traceability, and environmental sustainability serves as a multifaceted lens through which the AI ecosystem can be viewed, assessed, and improved. Our discussions have revealed that while challenges in balancing these principles persist, synergies also exist, offering opportunities for holistic advancements in AI regulation and application. As we look forward, it is clear that the development of informed, effective, and equitable AI regulatory systems will require ongoing effort, adaptation, and commitment to the principles outlined in the AI Act. This review serves as a foundational resource for policymakers, stakeholders, developers and scholars, aiming to navigate the complex regulatory landscape of AI and contribute to the development of informed, effective, and equitable regulatory AI systems.



\vskip 0.2in
\bibliography{main_ref}
\bibliographystyle{theapa}

\end{document}